%% file: nhien.tex
\documentclass[12pt]{article}

\input preamble.tex

\title{Darknet Traffic Classification and Adversarial Attacks}

\author{Nhien Rust-Nguyen\footnotemark[1]\ \ \ 
Mark Stamp\footnotemark[1]\,\,\footnotemark[2]}

\begin{document}

\symbolfootnotetext[1]{Department of Computer Science, San Jose State University}
\symbolfootnotetext[2]{mark.stamp$@$sjsu.edu}

\maketitle

\abstract
The anonymous nature of darknets is commonly exploited for illegal activities. Previous research has employed machine learning and deep learning techniques to automate the detection of darknet traffic in an attempt to block these criminal activities. This research aims to improve darknet traffic detection by assessing Support Vector Machines (SVM), Random Forest (RF), Convolutional Neural Networks (CNN), and Auxiliary-Classifier Generative Adversarial Networks (AC-GAN) for classification of such traffic and the underlying application types. We find that our RF model outperforms the state-of-the-art machine learning techniques used in prior work with the CIC-Darknet2020 dataset. To evaluate the robustness of our RF classifier, we obfuscate select application type classes to simulate realistic adversarial attack scenarios. We demonstrate that our best-performing classifier can be defeated by such attacks,  
and we consider ways to deal with such adversarial attacks.

\section{Introduction}

\begin{figure}[!htb]
	\centering
	\includegraphics[width=0.45\textwidth]{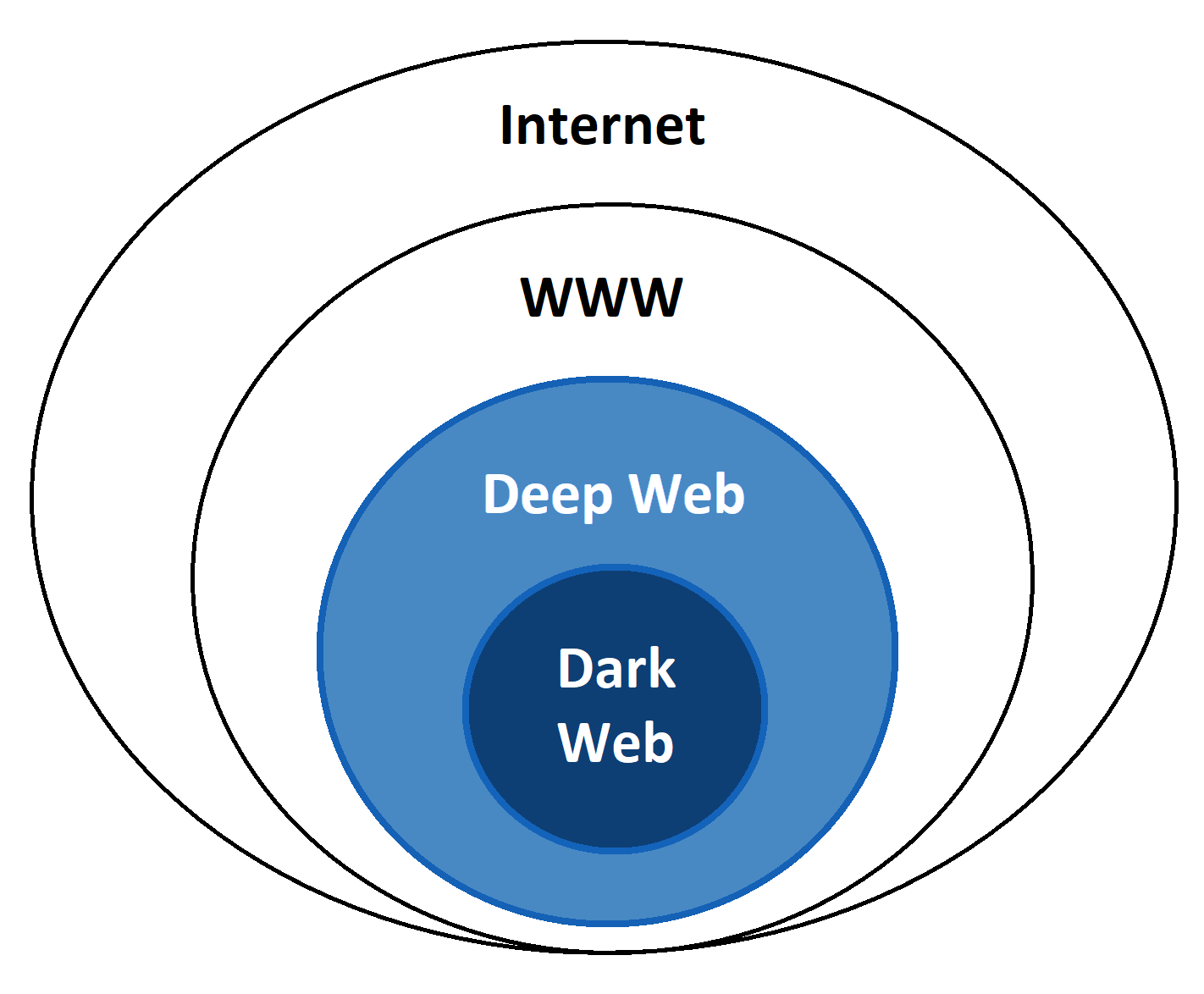}
	\caption{Layers of the Internet~\cite{wnn2021}} 
	\label{fig:web}
\end{figure}

Most of us are familiar with the Internet and the World Wide Web (WWW, or web). We regularly access both using web browsers or other networked applications to share information publicly, guided by search engine indexing of the Domain Name System (DNS) over globally bridged Internet Protocol (IP) networks. This publicly accessible and indexed address space is known as the surface web or clearnet. In contrast, the WWW address space which is not indexed by search engines but still publicly accessible is known as the deep web. Private networks within the deep web or networks comprised of unallocated address space are known as darknets and collectively termed the dark web. Figure~\ref{fig:web} illustrates the relationship between these layers of the Internet. 

The dark web is reached by an overlay network requiring special software, user authorization, or non-standard communication protocols~\cite{wnn2021}. Many darknets afford users anonymity during communication and thus facilitate many criminal activities, including hacking, media piracy, terrorism, trading illegal goods, human trafficking, and child pornography~\cite{gwern, cnnlstm2021}. 
Researchers are illuminating darknet traffic with machine learning and deep learning techniques, to better identify and inhibit these criminal activities. This research strives to contribute by promoting accurate classification of traffic features from the well-studied CIC-Darknet2020~\cite{didarknet} dataset. CIC-Darknet2020 is a collection of traffic features from two darknets, namely The Onion Router (Tor) and a Virtual Private Network (VPN), and also equivalent traffic generated over clearnet sessions using the same applications. We use classic machine learning techniques, such as Support Vector Machines (SVM) and Random Forest (RF) for classification of the network and application type. We also represent traffic features as grayscale images and apply deep learning architectures as classifiers, including Convolutional Neural Networks (CNN) and Auxiliary-Classifier Generative Adversarial Networks (AC-GAN). To assess the issue of extreme class imbalance within CIC-Darknet2020, we explore the option of dataset augmentation, assessing the Synthetic Minority Oversampling Technique (SMOTE) on the performance of our classifiers. Our results show that RF is the most effective at classifying both traffic type and the underlying application types.

Having established baseline classification performance, we consider the confusion of our best classifier in some detail, approaching the problem of darknet traffic detection adversarially. From the perspective of an attacker, we obfuscate the application classes in an attempt to avoid detection. We apply an encoding scheme to transform class features using probability analysis of the CIC-Darknet2020 dataset as a proof-of-concept. We strongly correlate the resulting RF confusion with our obfuscation technique for two attack scenarios, assuming few realistic limitations for traffic modification. We then assess the strength of our obfuscation technique with one defense scenario, by which we demonstrate that we can restore the performance of the RF classifier despite duress. We find that sufficient statistical knowledge of network traffic features can empower either classification or obfuscation tasks.

The rest of this paper is structured as follows. Section~\ref{chap:background} gives a brief background on Tor and VPN and discusses related work on darknet traffic detection. Section~\ref{chap:methodology} describes the dataset and outlines the methods employed for the experiments. Section~\ref{chap:implementation} provides background knowledge on the machine learning techniques used in our experiments and gives implementation details. Section~\ref{chap:results} discusses the results of our experiments. Lastly, Section~\ref{chap:conclusion} summarizes our research and explores possible future work.

\section{Background\label{chap:background}}

In this section, we first discuss the two broad categories of data in our dataset, namely, 
Tor and VPN traffic. Then we discuss the most relevant examples of related work.

\subsection{The Onion Router}
Initially, The Onion Router (Tor) was a project started by the United States Navy to secure government communication. Since 2006, Tor has become a nonprofit with thousands of servers (called relays or relay nodes) run by volunteers across the world~\cite{torproj}. Tor clients anonymize their TCP application IP addresses and sessions keys, sending encrypted application traffic through a network of relays~\cite{sarkar2020tor}. An example client application is the Tor Browser, which allows users to browse the web anonymously. 
Tor generally selects a relay path of three or more nodes and encrypts the data once for each node using temporary symmetric keys. The encrypted data hops from relay to relay, where each relay node only knows about the previous node and the next node along the path. This design makes it difficult to trace the original identity of Tor clients. Each relay removes a layer of encryption, so that by the last relay, the original data is forwarded to the intended destination as plaintext. Tor then deletes the temporary session keys used for encryption at each node, so that any compromised nodes cannot decrypt old traffic~\cite{dingledine2004tor}. 


\subsection{Virtual Private Networks}
Virtual Private Networks (VPN) are used to ensure communication privacy for individuals or enterprises, and can serve to separate private address spaces from the public Internet. VPN software disguises client IP addresses by tunneling encrypted communications through a trusted server, which acts as a gateway or proxy by routing client traffic to the broader network space. Client data is anonymized behind VPN server credentials before being forwarded to an intended destination, which may be either public or private. Any response traffic is sent back through the VPN server over the encrypted connection for the client to decrypt, ensuring anonymity between the client and recipient. Third parties, such as Internet Service Providers (ISP), will only see the VPN server as the destination of client communications. There are many forms of VPN. Some operate in the network layer, others reside in the transport or application layer~\cite{vpn}. 

\subsection{Related Work}
Many researchers have focused on the problem of detecting darknet traffic. However, there are limited public darknet datasets available. The CIC-Darknet2020 dataset used in the experiments reported in this paper was generated by~\cite{didarknet}. This dataset was also used in prior research, including~\cite{wnn2021, iliadis2021, cnnlstm2021}. It has become a well-known darknet traffic dataset due to its accessibility. In their research,~\cite{didarknet} grouped Tor and VPN together as darknet traffic, while non-Tor and non-VPN were grouped as benign traffic (clearnet). They created~$8\times 8$ grayscale images from~61 select features and used CNN to do classification on the dataset. Their CNN model achieved an overall accuracy of~94\% classifying traffic as darknet or benign and~86\% accuracy classifying the application type used to generate the traffic. The application traffic was broadly labeled as browsing, chat, email, file transfer, P2P, audio streaming, video streaming or VOIP.

Comparatively,~\cite{cnnlstm2021} experimented with classifying traffic and application type by combining a CNN and two other deep-learning techniques: Long Short-Term Memory (LSTM) and Gated Recurrent Units (GRU). They addressed the issue of having an imbalanced dataset by performing Synthetic Minority Oversampling Technique (SMOTE) on Tor, the minority traffic class. They used Principle Component Analysis (PCA), Decision Trees (DT), and Extreme Gradient Boosting (XGB) to extract~20 features before feeding the data into CNN-LSTM and CNN-GRU architectures. Their CNN layer was used to extract features from the input data, while LSTM and GRU did sequence prediction on these features. CNN-LSTM in combination with XGB as the feature selector produced the best F1-scores, achieving~96\% classifying traffic type and~89\% classifying application type.

The study~\cite{iliadis2021} focused on just traffic type from the CIC-Darknet2020 dataset. They used k-Nearest Neighbors (kNN), Multi-layer Perceptron (MLP), RF, DT, and Gradient Boosting (GB) to do binary and multi-class classification. For binary classification, they grouped the data into two classes (benign and darknet), similar to~\cite{didarknet}. For the multi-class problem, they used the original four classes of traffic type (Tor, non-Tor, VPN or non-VPN). They found that RF was the most proficient classifier for traffic type, yielding F1-scores of~98.7\% for binary classification and~98.61\% for multi-class classification.

Using the same dataset,~\cite{wnn2021} further broke down the application categories into~11 classes and used Weighted Agnostic Neural Networks (WANN) to classify the data. They proposed automating the tedious design of Artificial Neural Networks (ANN). Unlike regular ANN, WANN do not update neuron weights, but rather update their own network architecture piece-wise. WANN ranks different architectures by performance and complexity, forming new network layers from the highest ranked architecture. Their best WANN model achieved~92.68\% accuracy on application layer classification.

The UNB-CIC Tor and non-Tor dataset, also known as ISCXTor2016~\cite{tornontor2017}, was used by~\cite{sarkar2020tor} to classify Tor and non-Tor traffic using Deep Neural Networks (DNN). They built 2 models, \hbox{DNN-A} with 3-layers and \hbox{DNN-B} with 5-layers. \hbox{DNN-A} classified Tor from non-Tor samples with~98.81\% accuracy, while \hbox{DNN-B} achieved~99.89\% accuracy. For Tor samples, they built a 4-layer DNN to classify 8 application types. This model attained~95.6\% accuracy. 

In another study,~\cite{hu2021} generated their own dataset, capturing darknet traffic across 8 application categories (browsing, chat, email, file transfer, P2P, audio, video and VOIP) sourced from 4 different darknets (Tor, I2P, ZeroNet, and Freenet). They used a 3-layer hierarchical approach for classification. The first layer classified traffic as either darknet or normal. In the second layer, samples classified correctly as darknet were then classified by their darknet source. The third layer then classified application type for each of the darknet sources. The techniques~\cite{hu2021} used for classification include Logistic Regression (LR), RF, MLP, Gradient Boosting Decision Tree (GBDT), Light Gradient Boosting (LightGB), XGB, LSTM, and DT. Their hierarchical method attained~99.42\% accuracy in the first layer, 96.85\% accuracy in the second layer and~92.46\% accuracy in the third layer. Table~\ref{tab:prior} provides a summary of the prior work presented in this section.

\begin{table}[!htb]
	\caption{Summary of previous work}\label{tab:prior}
	\centering
	\adjustbox{scale=0.6}{
		\begin{tabular}{l|l|l|l|l} \midrule\midrule
			Work & Dataset & Problem considered & Techniques & Results \\
			\midrule
		\multirow{2}{*}{Demertizis, et al.~\cite{wnn2021}} & 
			\multirow{2}{*}{CIC-Darknet2020} & Only examines & \multirow{2}{*}{WANN} & \multirow{2}{*}{92.68\% accuracy}  \\
			    && 11 application types && \\ \midrule			
		\multirow{6}{*}{Hu, et al.~\cite{hu2021}} & 
			\multirow{6}{*}{Self-generated} & \multirow{2}{*}{Hierarchial approach:} & \multirow{3}{*}{LR, RF,} & Layer 1: \\ 
                		&& \multirow{2}{*}{Layer 1: darknet vs clearnet} & \multirow{3}{*}{MLP, GBDT,} & 99.42\% accuracy \\   
                		&& \multirow{2}{*}{Layer 2: Tor, I2P, ZeroNET}  &  \multirow{3}{*}{LightGB, XGB,} & Layer  2: \\ 
                		&& \multirow{2}{*}{\phantom{Layer 2: } and FreeNET} & \multirow{3}{*}{LSTM, DT} & 96.85\% accuracy  \\   
                		&& \multirow{2}{*}{Layer 3: 8 application types} && Layer 3: \\
                		&&&& 92.46\% accuracy  \\ \midrule
            	\multirow{4}{*}{Iliadis, et al.~\cite{iliadis2021}} & 
            		\multirow{4}{*}{CIC-Darknet2020} & \multirow{2}{*}{Only examines traffic type} & \multirow{3}{*}{kNN, MLP} & Binary: \\
                		&& \multirow{2}{*}{Binary: darknet vs clearnet} & \multirow{3}{*}{RF, DT, GB} & 98.7\% F1-score \\
                		&& \multirow{2}{*}{Multiclass: 4 traffic types} && Multiclass: \\                
                		&&&& 98.61\% F1-score \\ \midrule
		\multirow{4}{*}{Lashkari, et al.~\cite{didarknet}} & 
			\multirow{4}{*}{CIC-Darknet2020} & \multirow{3}{*}{Binary: darknet vs clearnet} & \multirow{4}{*}{CNN} & Binary: \\
			&& \multirow{3}{*}{Multiclass: 8 application types} && 94\% accuracy \\
			&&&& Multiclass: \\
			&&&& 86\% accuracy \\ \midrule
		\multirow{4}{*}{Sarkar, et al.~\cite{sarkar2020tor}} & 
			\multirow{4}{*}{ISCXTor2016} 
			& \multirow{2}{*}{Binary: Tor vs non-Tor} & \multirow{4}{*}{DNN}  & Binary: \\
                		&& \multirow{2}{*}{Multiclass:} && 99.89\% accuracy  \\ 
                		&& \multirow{2}{*}{8 application types within Tor} && Multiclass: \\
			&&&& 95.6\% accuracy  \\ \midrule
	        \multirow{4}{*}{Sarwar, et al.~\cite{cnnlstm2021}}  & 
	        		\multirow{4}{*}{CIC-Darknet2020}    & \multirow{3}{*}{4 traffic types} & \multirow{3}{*}{CNN-LSTM,} & Traffic: \\
	            	&& \multirow{3}{*}{8 application types} & \multirow{3}{*}{CNN-GRU}  
				& 96\% F1-score \\
			&&&& Application: \\
			&&&& 89\% F1-score \\ \midrule\midrule
		\end{tabular}
		}
\end{table}

\section{Methodology\label{chap:methodology}}

A primary goal of this research is to improve upon the state-of-the-art classification of darknet traffic by exploring the performance of SVM, RF, CNN, and AC-GAN as classifiers. We experiment with different levels of SMOTE during a preprocessing phase, oversampling the minority classes of the CIC-Darknet2020 dataset to assess the effects of data augmentation and class balance on classifier performance. We also experiment with representations of the darknet traffic features as 2-dimensional grayscale images for CNN and AC-GAN. Then we test the robustness of our best-performing classifier through a few obfuscation scenarios which serve to simulate adversarial attacks, assuming both the perspectives of an attacker and defender.

We apply statistical knowledge of the dataset to obfuscate our data features, disguising one or more classes as others. We explore three scenarios whereby we either obfuscate the training data, the test data or both. Obfuscating just the test data simulates an attack scenario in which traffic data is disguised while our classifier is yet unaware of the attack, and thus we can only apply previously trained models without a chance to learn of the obfuscation. Obfuscating just the training data simulates a scenario in which an attacker has accessed our training data to poison it, such that we train our classifier with malformed assumptions or outright malicious supervision. A third scenario supposes we collect some of the obfuscated traffic data before training our classifier, and thus have a chance to update our classification models to detect obfuscated test data.

\subsection{Dataset}
The CIC-Darknet2020 dataset~\cite{didarknet} is an amalgamation of two public datasets from the University of New Brunswick. It combines the ISCXTor2016 and ISCXVPN2016 datasets, which capture real-time traffic using Wireshark and TCPdump~\cite{tornontor2017, vpnnonvpn2016}. CICFlowMeter~\cite{cicflowmeter} is used to generate CIC-Darknet2020 dataset features from these traffic samples. Each CIC-Darknet2020 sample consists of traffic features extracted in this manner from raw traffic packet capture sessions. CIC-Darknet2020 consists of 158,659 hierarchically labeled samples. The top level traffic category labels consist of Tor, non-Tor, VPN or non-VPN. Within those categories, samples are further categorized by the types of applications used to generate the traffic. These subcategories include audio-streaming, browsing, chat, email, file transfer, P2P, video-streaming and VOIP. Table~\ref{tab:dataset} details the applications that are used to generate each type of traffic at the application level.

\begin{table}[!htb]
	\caption{CIC-Darknet2020 application classes~\cite{didarknet}}\label{tab:dataset}
	\centering
	\adjustbox{scale=0.85}{
		\begin{tabular}{l|l}\midrule\midrule
			Application class & Applications considered \\ \midrule
			Audio-Streaming & Vimeo and YouTube \\
			Browsing & Firefox and Chrome \\
			Chat & ICQ, AIM, Skype, Facebook and Hangouts \\
			Email & SMTPS, POP3S and IMAPS \\
			File Transfer & Skype and FileZilla \\
			P2P & uTorrent and Transmission (BitTorrent) \\
			Video-Streaming & Vimeo and YouTube \\
			VOIP & Facebook, Skype and Hangouts \\ \midrule\midrule
		\end{tabular}
		}
\end{table}

\subsection{Preprocessing}
The CIC-Darknet2020 dataset has samples with missing data, more specifically feature values of \texttt{NaN}. We remove samples with these values to improve the performance of the classifiers. As shown in Table~\ref{tab:trafsamples}, there are significantly less Tor samples compared to the other traffic categories. Prior work using this dataset eliminated CICFlowMeter flow labels, including: {\tt Flow Id}, {\tt Timestamp}, {\tt Source IP} and {\tt Destination IP}. The {\tt Flow Id} and {\tt Timestamp} are eliminated for this research as well. However, to retain as much information as possible from the CIC-Darknet2020 dataset, we separate each octet of the source and destination IP addresses into their own feature columns. Preliminary tests run on the dataset with and without these IP octet features indicate an improvement in the performance of the classifiers when the 
IP information is retained. 
Thus our feature dataset contains~72 features total after this preprocessing step.

\begin{table}[!htb]
\caption{Samples per traffic category}\label{tab:trafsamples}
	\centering 
	\adjustbox{scale=0.85}{
		\begin{tabular}{l|r}\midrule\midrule
		Traffic Category & Samples\\ \midrule
		Non-Tor & 93,357\zz \\
		Non-VPN & 23,864\zz \\
		Tor & 1,393\zz \\
		VPN & 22,920\zz \\ \midrule\midrule
		\end{tabular}
		}
\end{table}

The CIC-Darknet2020 dataset was scaled by min-max normalization, which applies the equation 
$$ 
  \mbox{{\tt normalizedValue}} = \frac{(\mbox{{\tt value}} - \mbox{{\tt min}})}{(\mbox{{\tt max}} - \mbox{{\tt min}})}
$$ 
to every value in each feature column to scale the feature values between~0 and~1. We also apply min-max normalization to our new IP octet feature columns.

\subsubsection{Data Balancing}
The CIC-Darknet2020 dataset does not have balanced sample counts among traffic and application classes, as shown in Tables~\ref{tab:trafsamples} and~\ref{tab:appsamples}. To explore the effect of reducing this imbalance on the classification task, we oversample each minority class using SMOTE. SMOTE interpolates linearly between feature values to produce new samples~\cite{bhagat2015smote}. We experiment with different levels of oversampling 
including~0\% (no SMOTE), 20\%, 40\%, 60\%, 80\% (partial SMOTE), and~100\% (full SMOTE). SMOTE is performed on all classes with less than the oversampling threshold as compared to the class with the largest sample count. Note that~100\% SMOTE results in an equal number of samples for each class, while lower thresholds of SMOTE result in an equal number of samples among only those classes which are oversampled.

\begin{table}[!htb]
\caption{Samples per application category}\label{tab:appsamples}
	\centering 
	\adjustbox{scale=0.85}{
		\begin{tabular}{l|r}\midrule\midrule
			Application Category & Samples \\ \midrule
			Audio-Streaming & 18,065\zz \\
			Browsing & 32,809\zz \\
			Chat & 11,479\zz \\
			Email & 6,146\zz \\
			File Transfer & 11,183\zz \\
			P2P & 48,521\zz \\
			Video-Streaming & 9,768\zz \\
			VOIP & 3,567\zz \\ \midrule\midrule
		\end{tabular}
		}
\end{table}

\subsubsection{Data Representation}
\begin{figure}[!htb]
	\centering
	\includegraphics[width=0.7\textwidth]{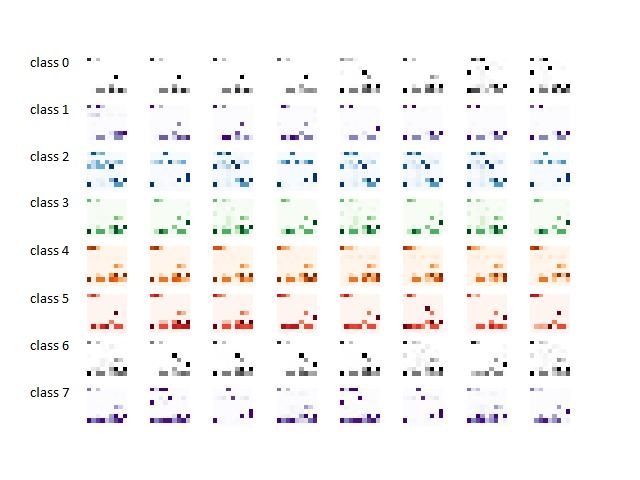}
	\vglue-20pt
	\caption{Data as 2-D images in original order}
	\label{fig:original}
\end{figure}

SVM and RF both use each dataset sample in its original format, which is a 1-dimensional array. However, we reshape each sample to be 2-dimensional for CNN and AC-GAN. Intuitively, the data is reshaped as~$9\times9$ grayscale images, where each of our 72 features is represented as a single pixel with the remaining pixels produced by zero padding. The pixels are ordered as their respective features appeared in the CIC-Darknet2020 dataset, starting at the top left corner of the image as shown in Figure~\ref{fig:original},
where each row represents samples from an application class, color-coded for readability.

\begin{figure}[!htb]
	\centering
	\includegraphics[width=0.7\textwidth]{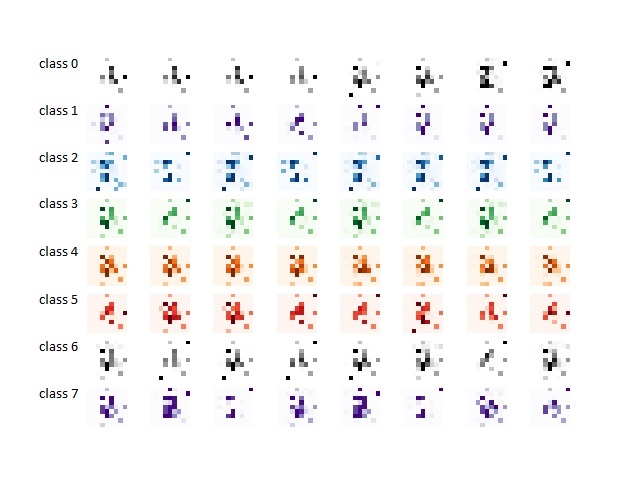}
	\vglue-20pt
	\caption{Data in 2-D sorted by RF feature importance and centered} 
	\label{fig:datarepRF}
\end{figure}

Both CNN and AC-GAN convolve local structures within the 2-D images, so adjacent pixels 
play an important role in classification. Therefore, we experiment with strategies to reorder the data to achieve better performance. We order the pixels by RF feature importance, starting at the top left corner of the image, and also reorganize the pixels spiraling outward from the center of the image. This latter strategy tends to group pixels with larger values toward the center of each image, as shown in Figure~\ref{fig:datarepRF}.

\subsubsection{Data Augmentation Experiment}
\begin{figure}[!htb]
	\centering
	\includegraphics[width=0.7\textwidth]{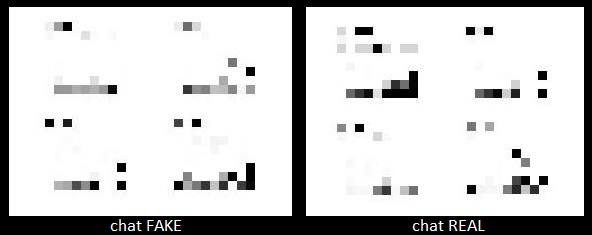}
	\caption{Chat class (4 fake and 4 real examples)} 
	\label{fig:fake}
\end{figure}

We experimented with AC-GAN as an alternative to SMOTE, to generate realistic artificial samples in an attempt to augment our dataset as another way to address the issue of class imbalance. However, we abandoned this approach as we found that the fake images generated by AC-GAN are consistently detectable by a CNN model with near perfect accuracy, 
ranging from~99\% to~100\% accuracy within a few epochs of training. The depth of our neural network architecture for AC-GAN was constrained by the input image size of our data. Therefore, we were unsuccessful in our attempt to use AC-GAN to augment our data. An example of some fake samples compared to real samples can be found in Figure~\ref{fig:fake}.

\subsection{Evaluation Metrics}
In our experiments, we use
accuracy and F1-score to measure the performance of each classifier. Accuracy is computed as the total number of correct predictions over the number of samples tested. The F1-score is the weighted average of precision and recall metrics, which is better for unbalanced datasets like CIC-Darknet2020. Similar to accuracy, F1-scores fall between~0 and~1, with 1 being the best possible score. F1-score is computed as 
$$ 
  \mbox{{\tt F1}} = {\tt 2} \times \frac{(\mbox{{\tt Precision}} \times \mbox{{\tt Recall}})}{(\mbox{{\tt Precision}} + \mbox{{\tt Recall}})}
$$ 
Precision calculates the ratio of samples classified correctly for the positive class, while recall measures the total number of positive samples that were classified correctly. Precision and recall are computed as
$$ 
  \mbox{{\tt Precision}} = \frac{\mbox{{\tt True Positives}}}{(\mbox{{\tt True Positives}} + \mbox{{\tt False Positives}})}
$$ 
and
$$ 
  \mbox{{\tt Recall}} = \frac{\mbox{{\tt True Positives}}}{(\mbox{{\tt True Positives}} + \mbox{{\tt False Negatives}})}
$$

\section{Implementation\label{chap:implementation}}

This section details the implementation of the experiments that
we mentioned in Section~\ref{chap:methodology}.
All experiments are coded in Python. The {\tt Imblearn} library is used to implement SMOTE to balance the dataset, while the package {\tt Scikit-learn} is employed to run SVM and RF experiments. The {\tt Tensorflow} and {\tt Keras} libraries are utilized to implement CNN and AC-GAN. From the {\tt Scikit-learn} library, the {\tt metrics} module is used to evaluate the F1-scores and accuracy of the classifiers and the {\tt StratifiedKFold} function is applied to perform 5-fold cross validation. Graphs are generated with the {\tt Matplotlib} and {\tt Seaborn} libraries. 

All experiments in this research are executed on one of two personal computers, as detailed in
Table~\ref{tab:computer}. We exploit a graphics processing unit (GPU) in the second computer to decrease the runtime duration of our more computationally demanding experiments, that is those using neural nets to process 2-D image representations. However, for portability and convenience we also use the first computer, which is a light-weight laptop.

\begin{table}[!htb]
	\caption{Computer hardware}\label{tab:computer}
	\centering
	\adjustbox{scale=0.85}{
		\begin{tabular}{l|l} \midrule\midrule
			Computer Processors & Experiments  \\
			\midrule
			\multirow{2}{*}{CPU: 8-core Intel(R) CORE(TM)}    & SVM \\
			\multirow{2}{*}{i7-8550U @ 1.80GHz}          & RF \\ 
			                            & Obfuscation \\ \midrule	
			CPU: 12-core Intel(R) Xeon(R)    & \multirow{2}{*}{CNN} \\  
			W-10855M @ 2.80GHz          & \multirow{2}{*}{AC-GAN}\\
			GPU: NVIDIA Quadro       & \multirow{2}{*}{SMOTE} \\
			RTX 5000 &\\ \midrule\midrule	
		\end{tabular}
		}
\end{table} 

\subsection{Background of Classification Techniques}

This section describes the machine learning and deep learning concepts that we
apply to classification in our experiments. These include SVM, RF, CNN, and AC-GAN.

\subsubsection{Support Vector Machines}
SVM are supervised machine learning models frequently used for classification. SVM find one or more hyperplanes to separate labeled training data while maximizing the margin of decision boundaries between classes. Maximizing the decision margin minimizes the classification error. The data must be vectorized into linear feature sets, but non-linear data can also be encoded with some success. Scaling the feature values across training samples allows coefficients of the hyperplanes (weights) to be ranked by relative importance. SVM rely on the kernel trick to map data into a higher dimensional space. The idea behind the kernel trick is that a larger feature space makes it easier to find hyperplanes between classes. Once hyperplanes are found, SVM can classify new samples into one of the classes as separated by the hyperplanes~\cite{stamp2018ml}. For our research, we perform preliminary tests to determine the best kernel for our dataset, with the result being the Gaussian radial basis function (RBF).



\subsubsection{Random Forest}
RF is an ensemble method that generalizes DT. While DT is a simple and efficient classification algorithm, it is sensitive to variance in the training data and prone to overfitting. RF compensates for these deficiencies by generating many subsets of the dataset, then randomly selecting features (with replacement) to train a DT for each subset. This process is called bootstrapping. To classify, RF takes the majority vote from all resulting DT in a process called aggregation. Together bootstrapping and aggregation is referred to as bagging~\cite{misra2020rf,stamp2018ml}. RF also produces a useful attribute called feature importance, which is a list of all the features ranked by the mean entropy within the DT. Feature importance tells us how influential each feature is when classifying samples with RF.


\subsubsection{Convolutional Neural Networks}
CNN are a unique type of neural network that operate on local structures, making them ideal for image analysis. CNN are composed of an image input layer, convolution and pooling layers and an output layer that produces a single vector of class scores. The neural nodes of convolution layers are connected to only a small portion of the nodes of the previous layer, with the exception of dense layers which have connections to every node. Convolution, pooling and dense layers are the fundamental components of any CNN architecture. In convolution layers, an input image is convolved with initially randomized filters to produce local structure maps that are joined to create the output of this layer. The filter windows slide across the input image, computing a dot product of the filter and image at each location. Pooling layers decrease total training time by reducing the dimensionality of the resulting feature maps, concentrating effort on the highest-value features~\cite{stanfordcnn,didarknet}. There are different types of pooling; for this research, we use max pooling. 

Our CNN architecture is based on that described in~\cite{didarknet}. We experiment with hyperparameters, running different combinations where we adjust the following.
\begin{itemize}
	\item Initial number of convolution filters (9, 32, 64, 81)
	\item Filter size ($2\times2$, $3\times3$)
	\item Percentage dropout $(0.2, 0.5)$
	\item Number of nodes in the first dense layer (72, 256)
\end{itemize}
All these architectures yield accuracies within the range of~86\% to~88\% when classifying application type. Therefore, we select the architecture that produces the highest accuracy. 
We use Adam for our optimizer and sparse categorical cross entropy for our loss function.



\begin{figure}[!htb]
	\centering
	\includegraphics[width=0.6\textwidth]{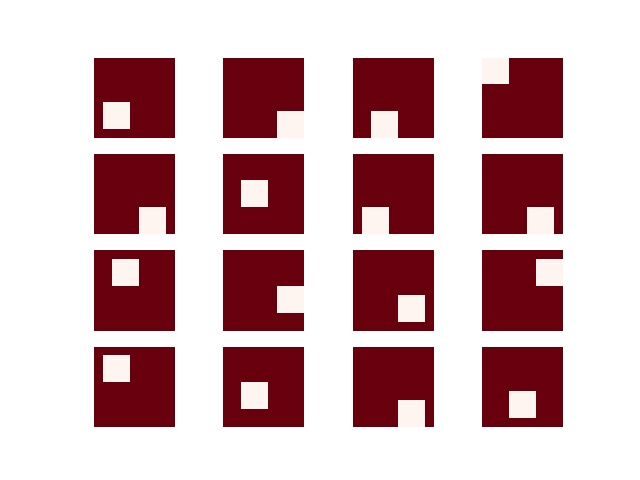}
	\vglue-10pt
	\caption{Examples $9\times9$ images with $3\times3$ cutouts} 
	\label{fig:cutout}
\end{figure}

Dropout is a common technique used in neural networks with fully-connected layers to combat overfitting. However, it is found to be not as effective with convolution layers. A better suited regularization technique for CNN is to cut out sections of the input images. Cutouts delete features to force CNN to learn from the other parts of each image during training. It is comparative in effect to dropout except that it operates on the input stage rather than the intermediate layers~\cite{cutouts2017, keystroke2021}. We implement cutouts by creating feature masks of equivalent size to our input image. We experiment with different cutout sizes including~$2\times2$, $3\times3$, and~$4\times4$ and randomize the position of the cutout within the mask. The masks are applied to our training data with matrix multiplication. Refer to Figure~\ref{fig:cutout} for some examples of masks with~$3\times3$ cutouts.

\subsubsection{Auxiliary-Classifier Generative Adversarial Network}
Generative Adversarial Networks (GAN) are comprised of two neural network architectures, a generator and a discriminator, that compete in a zero-sum game during training. The generator takes noise from a latent space as input and produces images that feed into the discriminator. The discriminator is given both real and generated images and is tasked to classify them as either real or fake. The discriminator error is then fed back into the generator as direction for image generation. AC-GAN is an extension of this base GAN architecture, taking a class label as additional input to the generator while predicting this label as part of the discriminator output. The objective of the AC-GAN generator is to minimize the ability of the discriminator to distinguish between real and fake images but also maximize the accuracy of the discriminator when predicting the class label~\cite{Mudavathu2018acgan,nagara2021acgan}. Besides using the AC-GAN generator in our data augmentation experiment, we also explore the secondary class prediction output of the discriminator as a classifier.


Our AC-GAN architecture is inspired by the ImageNet model described in~\cite{odena2017acgan}. However, since that architecture was built for image sizes~$32\times32$ and bigger, we modify that architecture to accommodate our~$9\times9$ image size by reducing the number of convolution and transposed convolution layers in the discriminator and generator, respectively. 

We fine-tune our hyperparameters by experimenting with the following.
\begin{itemize}
\item Latent space size (81, 100)
\item Initial number of convolution filters (15, 40, 64, 192, 202, 384, 500, 1500)
\item Number of nodes in the first dense layer (31, 81, 128, 384, 405, 768, 1000, 3000)
\item Filter size ($3\times3$, $5\times5$)
\item Stride size ($2\times2$, $3\times3$)
\end{itemize}
We observe accuracies within the range of~70\% to~73\% when classifying application type with these hyperparameters. The best-performing architecture with the shortest runtime duration is used in this research. Tables~\ref{tab:generator} and~\ref{tab:discriminator} detail our generator and discriminator architecture, respectively. 

We feed training data to our AC-GAN model in batches of 64 samples. Batch normalization (BN) layers are applied between convolution layers to regularize the training gradient step size. This is thought to smooth local optimization steps and stabilize training, thereby accelerating convergence of GAN models~\cite{batchnorm}.

\begin{table}[!htb]
	\caption{AC-GAN generator architecture}\label{tab:generator}
	\centering
	\adjustbox{scale=0.85}{
		\begin{tabular}{lcccccc} \midrule\midrule
			Layer Operation & Kernel & Strides & Depth & BN & Activation  \\
			\midrule
			$1\times 1\times 100$ Input A (Latent Space) &   &   &   &   &  \\
			Dense A &   &   & 405 &   & ReLU \\
			$1\times 1\times 1$ Input B (Feature Noise) &   &   &   &   &  \\
			$8\times 32$ Class Embedding for B &   &   & 256 &   &   \\
			Dense B &   &   & 1 &   &   \\
			Merge A + B &   &   & 406 &   &  \\
			Conv2DTranspose & $5\times 5$ & $3\times 3$ & 202 & \checkmark & ReLU \\
			Conv2DTranspose & $5\times 5$ & $3\times 3$ & 1 &   & Tanh \\ \midrule\midrule	
		\end{tabular}
	} 
\end{table}

\begin{table}[!htb]
	\caption{AC-GAN discriminator architecture}\label{tab:discriminator}
	\centering
	\adjustbox{scale=0.75}{
		\begin{tabular}{lcccccl} \midrule\midrule
			Layer Operation & Kernel & Strides & Depth & BN & Dropout & Activation  \\
			\midrule
			9x9x1 Input (Image) \\
			Conv2D & $3\times3$ & $2\times2$ & 32 &  & 0.5 & Leaky ReLU \\
			Conv2D & $3\times3$ & $1\times1$ & 64 & \checkmark & 0.5 & Leaky ReLU \\
			Conv2D & $3\times3$ & $2\times2$ & 128 & \checkmark & 0.5 & Leaky ReLU \\
			Conv2D & $3\times3$ & $1\times1$ & 256 & \checkmark & 0.5 & Leaky ReLU \\
			Flatten \\
			Dense & & & 1 & & & Sigmoid \\
			Dense & & & 8 & & & Softmax \\ \midrule
			Leaky ReLU Slope & & & & & & 0.2 \\
        	Weight Initialization & & & & & & Gaussian $(\sigma=0.02)$ \\
        	Optimizer & & & & & & Adam $(\alpha=0.0002,  \beta_1=0.5)$\\
			\midrule\midrule	
		\end{tabular}
	}
\end{table}

\subsection{Adversarial Attacks}
Our adversarial attacks rely on on obfuscation, which serves to disguise application classes 
based on applied probability analysis. We select 
application classes to disguise as other classes based on minimum and maximum sum statistical distance between all class features, as specified in Algorithm~\ref{tab:sumalg}. 

\begin{algorithm}[!htb]
\caption{Class feature probability distributions}\label{tab:sumalg}
 \begin{algorithmic}[1]
 \footnotesize
    \Procedure{compare}{$classA,classB$}
    \State bins = some discrete bins partitioning values 0 to 1
    \Comment{We use 100 bins.}
    \State A = classA feature probability distributions
    \State B = classB feature probability distributions
    \State classDistance = 0
    \For{\textbf{each} distributionA, distributionB in A, B}
        \State featureDistance = {\tt cdist}(distributionA, distributionB)
        \Comment{Euclidean}
        \State classDistance += featureDistance
        \Comment{Manhattan Sum}
    \EndFor
    \EndProcedure
\end{algorithmic}
\end{algorithm}

We also select a third class transformation to perform based on maximal classifier confusion, whose sum statistical distance between class features is notably low but not the minimum between classes. We ensure our class transformation can be decoded by encoding features with a deterministic algorithm, Algorithm~\ref{tab:obalg}, but impose no other restrictions on feature transformation.

\begin{algorithm}[!htb]
\caption{Disguise one class sample as another class sample}\label{tab:obalg}
\begin{algorithmic}[1]
\footnotesize
\Procedure{obfuscate}{$sample,classA,classB$}
    \Comment{To decode, reverse A and B}
    \State bins = some discrete bins partitioning values 0 to 1
    \Comment{We use 100 bins.}
    \State A = classA feature probability distributions
    \State B = classB feature probability distributions
    \For{\textbf{each} featureValue at featureIndex in the sample}
        \State featureBin = the bin which contains featureValue
        \State DCPD = A[featureIndex] - B[featureIndex]
        \State AtoB = sorted DCPD from maximum to minimum
        \State BtoA = sorted DCPD from minimum to maximum
        \State oldBin = where AtoB[oldBin] == featureValue
        \Comment{Red arrows in Figure~\ref{fig:dcpd}}
        \State newBin = BtoA[oldBin]
        \Comment{Black arrows in Figure~\ref{fig:dcpd}}
        \State newValue = featureValue - (bins[oldBin] - bins[newBin])
        \State sample[featureIndex] = newValue
        \EndFor
\EndProcedure
\end{algorithmic}
\end{algorithm}

\begin{figure}[!htb]
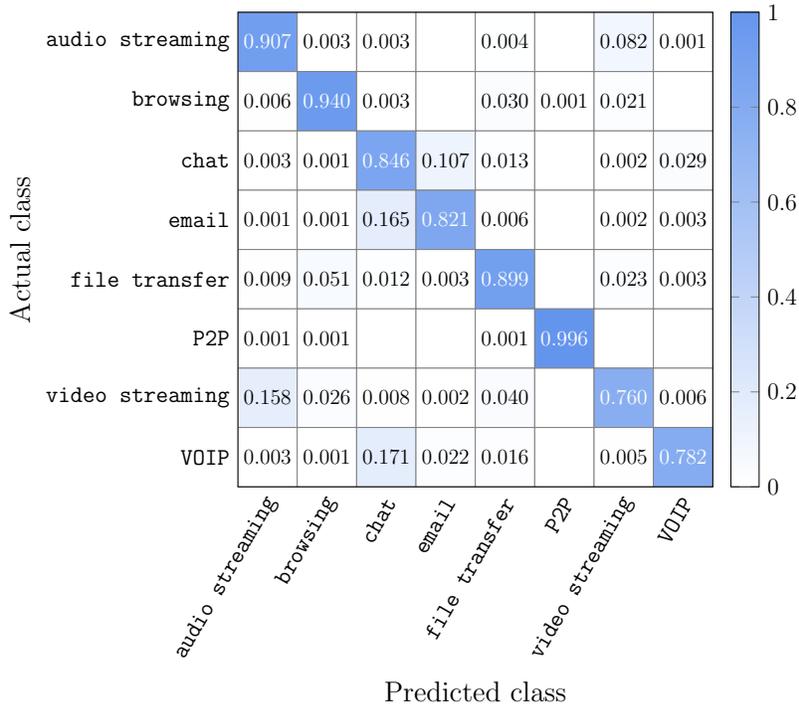

	\centering
	\input figures/conf_RF.tex
	\caption{Confusion matrix for best RF classifying level 2} 
	\label{fig:rf2}
\end{figure}

We start by generating normalized histograms of feature values per class to assess the probability at which values occur within each class over our entire dataset. To decide which classes to obfuscate, we examine the sums of the distances between feature probability distributions from each class to each other class, depicted as an~$8\times8$ table in the case of application type as provided in Table~\ref{tab:sum}. We use the {\tt cdist} function of the {\tt scipy} Python library to calculate the Euclidean distance between probability distributions. This allows us an estimation of the overall difference between classes while considering all feature probability distributions equally. We pick the classes with the minimum and maximum sum of statistical distances between features, changing class~0 (audiostreaming) to class~5 (P2P) and class~3 (email) to class~7 (VOIP). We also examine the confusion matrix for our best-performing classifier\textemdash RF, shown in Figure~\ref{fig:rf2}. RF is observed to be most confused between class~2 (chat) and class~3 (email), so we decide to additionally obfuscate class~2 with class~3. We arbitrarily choose to transform lower numbered classes to higher numbered classes (e.g. disguising class~2 as 
class~3 instead of class~3 as class~2).

\begin{table}[!htb]
	\caption{Statistical distances between pairs of application classes}\label{tab:sum}
	\centering
	\adjustbox{scale=0.85}{
		\begin{tabular}{l|c|c|c|c|c|c|c|c} \midrule\midrule
			Class & 0 & 1 & 2 & 3 & 4 & 5 & 6 & 7 \\
			\midrule
			0 & 0 &	25.129 & 18.709 & 21.041 & 23.656 &	28.195 \cellcolor[gray]{.9} & 18.371 & 21.903 \\ \midrule	
			1 & 25.129 & 0 & 23.518 & 21.958 & 12.884 & 12.098 & 16.728 & 23.623 \\ \midrule	
            2 & 18.709 & 23.518 & 0 & 9.841 \cellcolor[gray]{.75} & 22.613 & 25.294 & 18.408 & 9.901 \\ \midrule	
            3 & 21.041 & 21.958 & 9.841 \cellcolor[gray]{.75} & 0 & 21.51 & 23.021 & 18.031 & 6.859 \cellcolor[gray]{.6} \\ \midrule	
            4 & 23.656 & 12.884	& 22.613 & 21.51 & 0 & 15.651 & 14.605 & 23.211 \\ \midrule	
            5 & 28.195 \cellcolor[gray]{.9} & 12.098 & 25.294 & 23.021 & 15.651 & 0 & 21.085 & 24.451 \\ \midrule	
            6 & 18.371 & 16.728 & 18.408 & 18.031 & 14.605 & 21.085 & 0 & 20.089 \\ \midrule	
            7 & 21.903 & 23.623 & 9.901 & 6.859 \cellcolor[gray]{.6} & 23.211 & 24.451 & 20.089 & 0 \\ \midrule\midrule	
		\end{tabular}
		} 
\end{table} 


Our obfuscation algorithm first calculates the difference in class probability distributions (DCPD) for each feature between 2 classes, denoted as~A and~B, and sorts each distribution from maximum to minimum. Intuitively the index of each DCPD maximum corresponds to each feature value most probably belonging to the positive class~A while the minimum corresponds to each feature value most probably belonging to the negative class~B. To obfuscate a sample, we then transform individual feature values by subtracting the difference in bin thresholds between the original feature bin and a target bin for obfuscation. To choose target bins for this transformation, we create a 1-to-1 map of the sorted indices of each DCPD with a reverse sort of the same DCPD. This ensures a transformed sample feature could be decoded later given the feature DCPD for a class obfuscation vector. 

An example visualization of DCPD bin mapping for the transformation of the most 
common feature~0 values from class~2 to class~3 is provided in Figure~\ref{fig:dcpd}. Reversing the 1:1 bin map facilitates decoding of obfuscated class sample feature values back to their original values. To do this we add back the same difference in bin thresholds which we subtracted earlier, thus applying each feature DCPD between known classes as a decoder key to undo an expected class obfuscation for a particular feature. To test this method of class obfuscation, we performed the three experiments described in Section~\ref{chap:methodology} and summarized in Table~\ref{tab:obfexp}, with RF as the classifier.

\begin{table}[!htb]
\caption{Obfuscation scenarios}\label{tab:obfexp}
	\centering
	\adjustbox{scale=0.85}{
		\begin{tabular}{l|c|c|l}\midrule\midrule
		\multicolumn{1}{r|}{} & \multicolumn{2}{c|}{What is obfuscated?}  & \multirow{2}{*}{Scenario description}\\
		                       & Training data & Test data &  \\ \midrule
		\multirow{3}{*}{Case~1}           && \multirow{3}{*}{\checkmark}   & Simulates a novel attack \\
		\multirow{3}{*}{}                 && \multirow{3}{*}{}   & where we apply an outdated \\
		\multirow{3}{*}{}                 && \multirow{3}{*}{}   & model for classification \\ \midrule
		\multirow{3}{*}{Case~2}           & \multirow{3}{*}{\checkmark} && Simulates an attack \\
        \multirow{3}{*}{}                 && \multirow{3}{*}{}   & on our training data, \\
		\multirow{3}{*}{}                 && \multirow{3}{*}{}   & poisoning the classifier \\ \midrule
	    \multirow{3}{*}{Case 3}           & \multirow{3}{*}{\checkmark} & \multirow{3}{*}{\checkmark}   & Simulates a novel defense \\
		\multirow{3}{*}{}                 && \multirow{3}{*}{}   & where we train our model \\ 
		\multirow{3}{*}{}                 && \multirow{3}{*}{}   & on some obfuscated data \\ \midrule\midrule
		\end{tabular}
		}
\end{table}

\subsubsection{An Obfuscation Example}
\begin{figure}[!htb]
	\centering
	\includegraphics[width=0.8\textwidth]{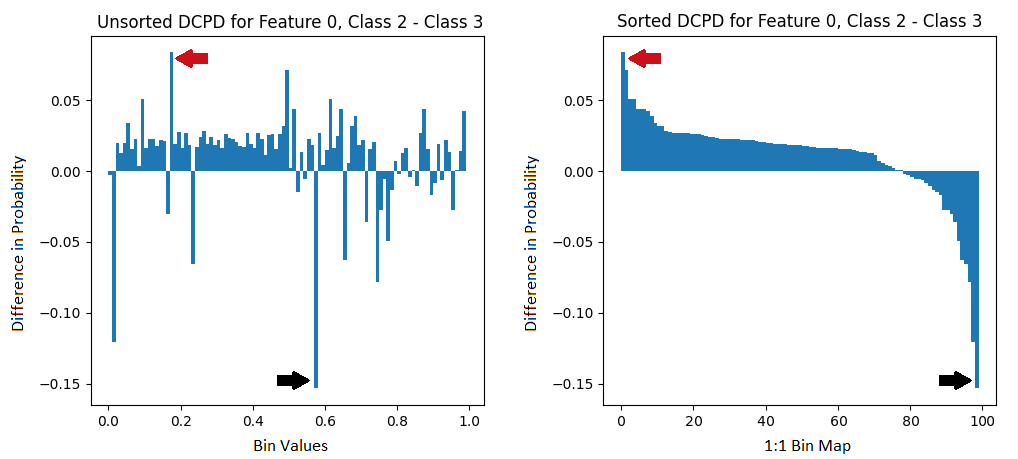}
	\caption{Visualization of obfuscation example}
	\label{fig:dcpd}
\end{figure}

To better illustrate Algorithm~\ref{tab:obalg}, we will walk through a simple example where we are given a sample from class~2 and we want to transform this sample to look like class~3. Let us start with the first feature, feature~0. We note the value of this feature for class~2; call this value~$v$. 
Suppose, for example, that~$v = 0.178142$. We allocate 100 equal-width bins ranging from~0 to~1, so that bin~$b_0$ corresponds to values~0.00 to~0.01 and so on. Given the value of~$v$, we find the bin that~$v$ falls into. The value~$v= 0.178142$ is in bin~$b_{17}$, which contains values between~0.17 to~0.18. Bin~$b_{17}$ is indicated by the red arrows in Figure~\ref{fig:dcpd}. We then flip the sorted DCPD index at~$b_{17}$ to locate our target bin, indicated by the black arrows in Figure~\ref{fig:dcpd}. This target bin~$b_{58}$, which contains values between~0.58 to~0.59. To obfuscate, we subtract the difference between~$b_{17}$ and~$b_{58}$ from~$v$. In this example, our new transformed value is
$$
  \widetilde{v} = 0.178142 - (0.17 - 0.58) = 0.588142
$$
which falls into the target bin~$b_{58}$. We repeat this for all the features to transform the sample from class~2 to class~3.

Note that this obfuscation technique is designed to maximize the effectiveness of
a simulated adversarial attack. Our approach ignores practical limitations on the ability
of attackers to modify the statistics of the data. Hence these simulated attacks 
can be considered worst-case scenarios, from the perspective of 
detecting darknet traffic under adversarial attack.

\section{Results and Discussion\label{chap:results}}

In this section, we consider a wide range of experiments. First, we determine
which of the three 2-D image representation techniques discussed in
Section~\ref{sect:dataRep} is most effective. Then we consider the use
of cutouts, which can serve to reduce overfitting and improve accuracy in CNNs.
We then turn our attention to the imbalance problem, with a series of SMOTE
experiments. We conclude this section with an extensive set of experiments
involving various adversarial attack scenarios.

\subsection{Data Representation Experiments}\label{sect:dataRep}
We evaluate CNN and the AC-GAN discriminator given different 2-D pixel representations of the data features. All of our 2-D representations of the data are of size~$9\times9$, where each pixel is a feature. The pixels in the original representation follow the order that the features appear in the CIC-Darknet2020 dataset. We hypothesize that grouping the pixels together would have a positive effect on the performance of our classifiers since convolutions operate on local structures. Our results show that CNN performs best when the pixels are sorted by RF feature importance and then grouped together at center of the image. However, this is not true for the AC-GAN discriminator. AC-GAN does better using the original data representation, contrary to our hypothesis. Table~\ref{tab:compdatarep} shows the results for these experiments.

\begin{table}[!htb]
	\caption{2-D data representation performance for application classification\label{tab:compdatarep}}
	\centering
	    \adjustbox{scale=0.75}{
		\begin{tabular}{l|cc|cc} \midrule\midrule
			\multicolumn{1}{r|}{} & \multicolumn{2}{c|}{CNN} & \multicolumn{2}{c}{AC-GAN} \\
			& Accuracy & F1-scores & Accuracy & F1-score \\
			\midrule
			Original & 0.889 & 0.887 & 0.753 & 0.738 \\
			Shaped with RF feature importance & 0.890 & 0.887 & 0.753 & 0.731 \\
			Shaped with RF feature importance \& centered & 0.891 & 0.889 & 0.742 & 0.729 \\ \midrule\midrule
		\end{tabular}
		}
\end{table}

\subsection{Cutout Experiments}
\begin{figure}[!htb]
	\centering
	\includegraphics[width=0.8\textwidth]{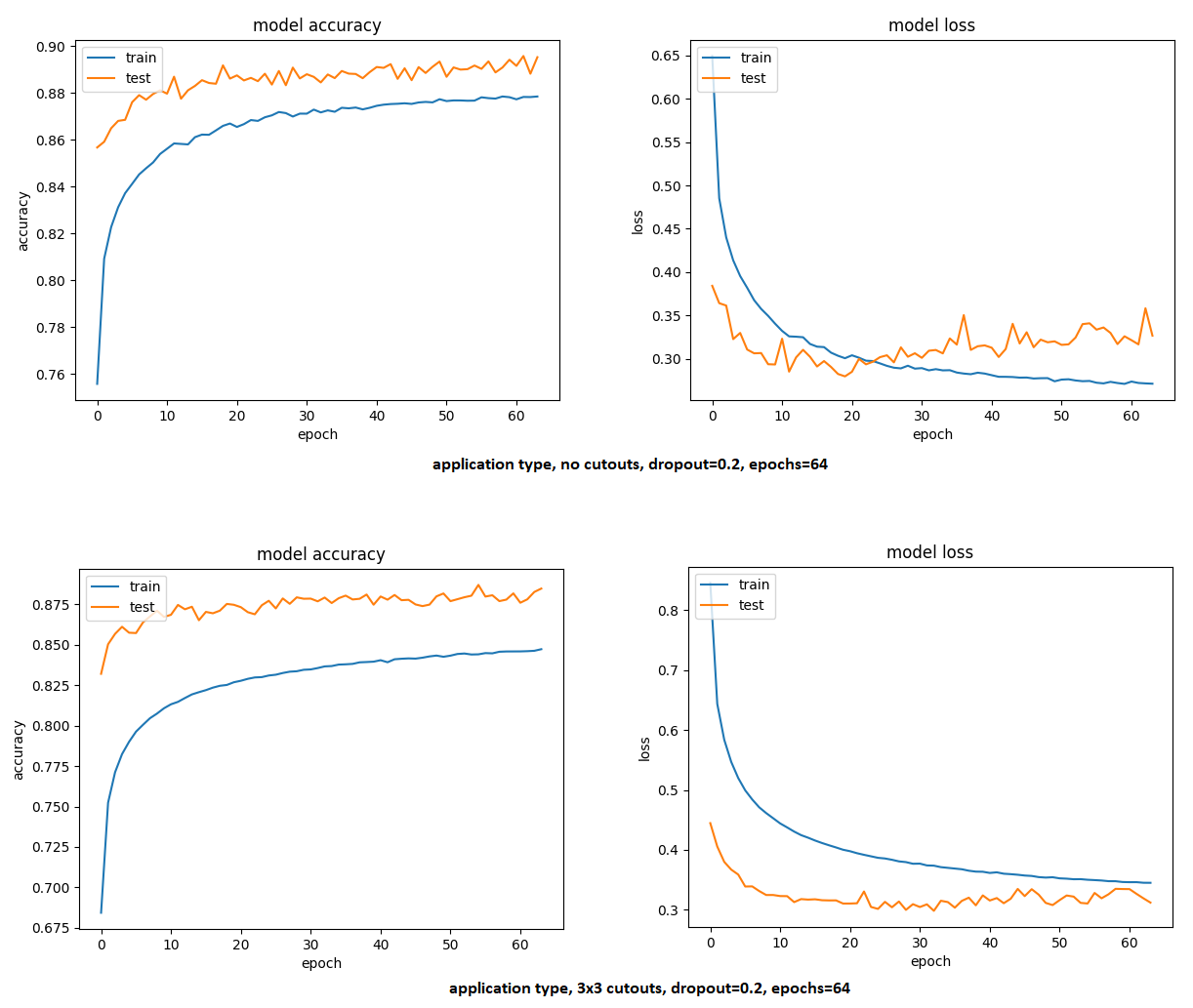}
	\caption{The effects of cutouts on overfitting for CNN} 
	\label{fig:cnncutout}
\end{figure}

Initially, our CNN model is able to achieve~88\% accuracy classifying application type within 15 epochs. However, we notice that overfitting starts to occur the longer we run our model. To reduce overfitting, we apply cutouts to the training data. We experiment with different cutout sizes: $2\times2$, $3\times3$, and~$4\times4$. We observe that cutouts allow our CNN to train for a longer period of time without overfitting. The loss graphs in Figure~\ref{fig:cnncutout} show how the CNN model overfits after 20 epochs in the original execution but does not overfit with cutouts. There is little difference in the effects of applying~$2\times2$ compared to~$3\times3$ cutouts. Both delay overfitting at the same rate and the accuracies for both linger at~88\%. Notably, we witness a~1\% decrease in accuracy with~$4\times4$ cutouts. As our images are only~$9\times9$ pixels, a~$4\times4$ cutout may delete too much information from the image, negatively affecting the accuracy. While cutouts address the issue of overfitting, we find that more training does not significantly improve the performance of CNN for our dataset.

\subsection{SMOTE Experiments}

We compare the performance of our classifiers with various levels of SMOTE, performing SMOTE to oversample the training data before training each classifier to classify traffic and application types; the results from these experiments appear in Tables~\ref{tab:smotelvl1}
and~\ref{tab:smotelvl2}, respectively.
We observe
that reducing class imbalance this way does not have a significant effect on the performance of RF and CNN. In contrast, we do see improved performance with SVM for both traffic and application classification tasks as the level of applied SMOTE increases. We also notice improved performance with increasing SMOTE level for AC-GAN when doing traffic type classification. We observe that partial SMOTE improves the performance of AC-GAN for application type classification, while noting that the performance decreases with full SMOTE. 

\begin{table}[!htb]
	\caption{Traffic classification F1-scores for increasing levels of SMOTE}\label{tab:smotelvl1}
	\centering
	\adjustbox{scale=0.85}{
		\begin{tabular}{r|cccc} \midrule\midrule
		SMOTE level & SVM & RF & CNN & AC-GAN \\
			\midrule
			0\%\zz\zz\zz\zz    & 0.986  & 0.998  & 0.998 & 0.974 \\ 
			20\%\zz\zz\zz\zz   & 0.993  & 0.998  & 0.995 & 0.980 \\
			40\%\zz\zz\zz\zz   & 0.993  & 0.998  & 0.995 & 0.984 \\
			60\%\zz\zz\zz\zz   & 0.993  & 0.998  & 0.995 & 0.986 \\
			80\%\zz\zz\zz\zz   & 0.993  & 0.998  & 0.996 & 0.987 \\
			100\%\zz\zz\zz\zz  & 0.993  & 0.998  & 0.995 & 0.987 \\ \midrule\midrule
		\end{tabular}
		}
\end{table}

\begin{table}[!htb]
	\caption{Application classification F1-scores for increasing levels of SMOTE}\label{tab:smotelvl2}
	\centering
	\adjustbox{scale=0.85}{
		\begin{tabular}{r|cccc} \midrule\midrule
		SMOTE level & SVM & RF & CNN & AC-GAN \\
			\midrule
			0\%\zz\zz\zz\zz    & 0.834  & 0.922  & 0.887 & 0.738 \\ 
			20\%\zz\zz\zz\zz   & 0.839  & 0.920  & 0.883 & 0.750 \\
			40\%\zz\zz\zz\zz   & 0.842  & 0.921  & 0.883 & 0.762 \\
			60\%\zz\zz\zz\zz   & 0.846  & 0.921  & 0.887 & 0.768 \\
			80\%\zz\zz\zz\zz   & 0.847  & 0.920  & 0.888 & 0.767 \\
			100\%\zz\zz\zz\zz  & 0.848  & 0.920  & 0.885 & 0.759 \\ \midrule\midrule
		\end{tabular}
		}
\end{table}

Our RF model without SMOTE outperforms the state-of-the-art F1-scores for both traffic and application classification tasks. We observe a~1.1\% improvement for traffic classification as compared to~\cite{iliadis2021}, where they also found RF to be their best classifier. Study~\cite{iliadis2021} only classified traffic type, thus no application type performance is available for comparison. For application classification, our RF model achieved a~3.2\% increase over~\cite{cnnlstm2021}. In addition, our CNN model outperformed
the CNN results in~\cite{didarknet} by~2.8\%. We are only able to compare classification results for application type with study~\cite{didarknet} because they approach traffic type classification as a binary problem while we address it as a multiclass problem. Table~\ref{tab:compprior} summarizes the best performance of our classifiers with relevant prior work. Overall, RF is our best-performing classifier and AC-GAN discriminator was shown to be our worst-performing classifier at any SMOTE level.

\begin{table}[!htb]
	\caption{Best F1-score compared to prior work}\label{tab:compprior}
	\centering
	\adjustbox{scale=0.8}{
		\begin{tabular}{l|ccc|cccc} \midrule\midrule
		\multirow{2}{*}{Data type}
		& \multicolumn{3}{c|}{Previous work} & \multicolumn{4}{c}{Our results}\\
			 & CNN-LSTM~\cite{cnnlstm2021} & RF~\cite{iliadis2021} & CNN~\cite{didarknet} & SVM & RF & CNN & AC-GAN  \\
			\midrule
			Traffic & 0.960 & 0.987 & --- & 0.993 & 0.998 & 0.998 & 0.987 \\
			Application & 0.890 & --- & 0.860 & 0.848 & 0.922 & 0.888 & 0.759 \\ \midrule\midrule			
		\end{tabular}
	}
\end{table}

\subsection{Adversarial Attack Experiments}
With continual improvement in the accuracy of darknet traffic detection by machine learning and deep learning techniques, it is realistic to anticipate that attackers will attempt to find ways to circumvent detection by modifying the profile of their application traffic. For example, someone pirating copyrighted media with P2P applications might disguise their illegal activity as VOIP traffic to avoid prosecution. We show obfuscation of traffic in this fashion can be accomplished by modifying traffic feature values, understanding that this process is most feasible and desirable at the application layer. Also, if an attacker were to discover the methods we use for classification and pollute our training data, then our classifiers could be compromised, allowing the attacker to avoid detection without modifying any of their traffic features.

For this experiment we assume the role of an attacker on the network, with the goal of modifying traffic features such that classes are incorrectly classified or entirely undetected. This could represent covert illegal activity that an attacker wishes to hinder the detection of, common examples being P2P or file-transfer applications. Realistically, traffic features common to one application class could be modified at the application layer to appear more similar to other application classes. An attacker could do this by writing a custom overlay application to change various features, such as the number of packets sent, their communication intervals, port assignment or many others. In our experiments, we disguise class~0 as class~5 (originally the most different), class~2 as class~3 (the classes which most confused our RF classifier) 
and class~3 as class~7 (originally the most similar).

In scenario case~1, we train our RF classifier on the original application class data, then test the same model with an obfuscated class in the test dataset. This represents a hypothetical scenario where an attacker modifies the traffic features of one class at the application layer, perhaps with an overlay application. We demonstrate that our method of obfuscation is able to defeat our best classifier in this scenario, significantly reducing detection of the obfuscated class, as well as overall classifier accuracy. Before obfuscation, RF classifies application classes with an accuracy of~92.3\% without SMOTE. After obfuscation of our three class choices mentioned in the previous paragraph, the overall RF accuracy decreases to~80.8\%, 85.4\%, and~88.7\% respectively, for application classification without SMOTE. 

\begin{figure}[!htb]
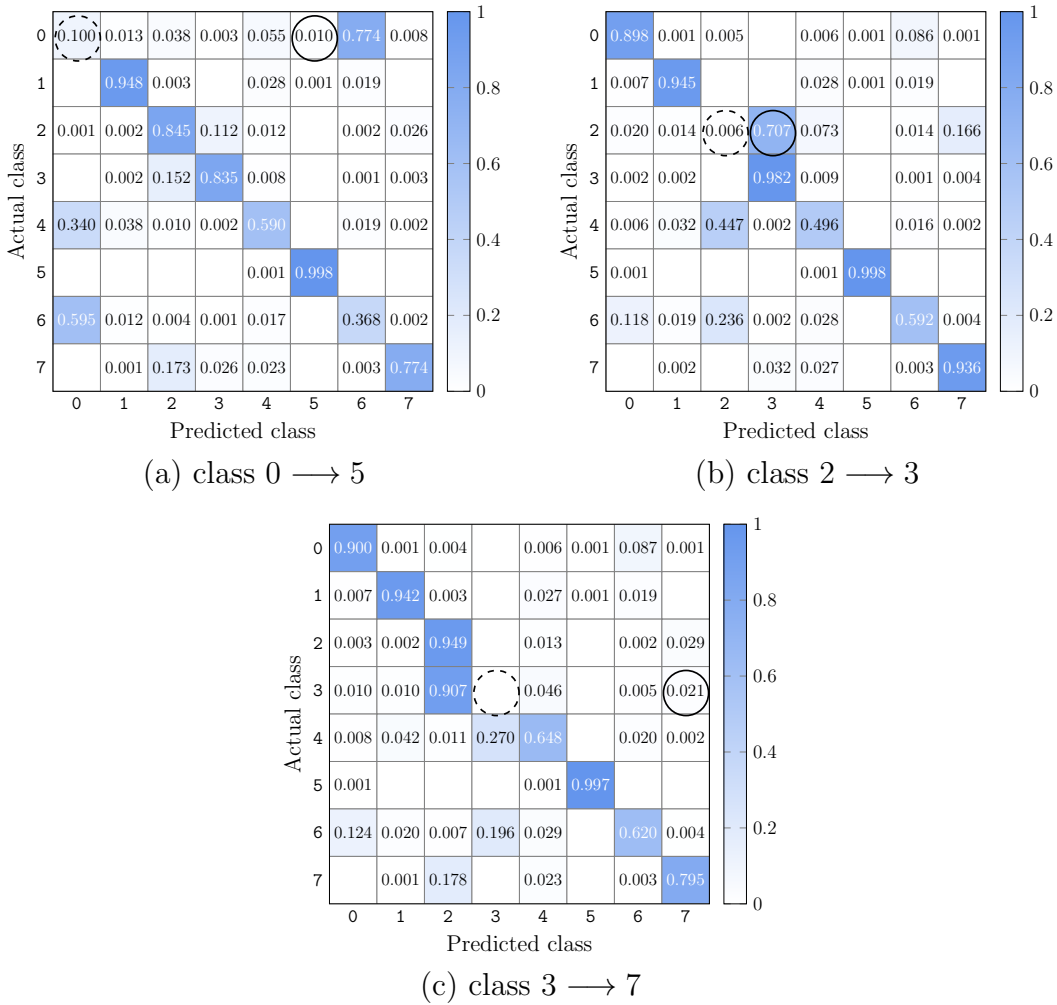

	\centering
	\begin{tabular}{cc}
	\adjustbox{scale=0.8}{
	\input figures/conf_1_a.tex
	}
	&
	\adjustbox{scale=0.8}{
	\input figures/conf_1_b.tex
	}
	\\
	(a) class $0 \longrightarrow 5$
	&
	(b) class $2 \longrightarrow 3$
	\\
	\\[-1.5ex]
	\multicolumn{2}{c}{\adjustbox{scale=0.8}{
	\input figures/conf_1_c.tex
	}}
	\\
	\multicolumn{2}{c}{(c) class $3 \longrightarrow 7$}
	\end{tabular}
	\caption{Confusion matrices for obfuscation scenario case~1} 
	\label{fig:case1}
\end{figure}

The confusion matrices in Figure~\ref{fig:case1} show that RF consistently misclassifies each class we obfuscate, actually detecting no samples in the case of an obfuscated class~3, where the dashed circle indicates the class we are obfuscating and the solid circle indicates the class we intended it to appear as.
However, RF did not misclassify classes~0 and~3 as the expected classes~5 and~7, respectively. Instead, the confusion matrices (a) and (c) in Figure~\ref{fig:case1} reveal that RF mostly categorizes class~0 and~3 as class~6 and~2, respectively. It may be relevant that our obfuscation method does not account for any interdependance between traffic feature values, obfuscating each feature independently.


In scenario case~2, we train our RF classifier with an obfuscated class in the training dataset, then test the model with the original application class data. We consider a hypothetical scenarios where an attacker entirely poisons our training data, perhaps by injecting malware into our database or by intercepting our traffic capture data stream. We find the attacker could prevent an entire class from being predicted by our best classifier when the training data for a class is entirely obfuscated. We see this trend in the all three confusion matrices in Figure~\ref{fig:case2},
where in each case, 	the dashed circle indicates the class we are obfuscating and the solid circle indicates the class we intended it to appear as.
Notice that the entire row in the confusion matrix is zeroed out, indicating that the class was never predicted for classification by RF. Similar to scenario case~1, the overall RF accuracy 
decreases to~82.0\%, 86.5\%, and~89.4\% respectively, for application classification without SMOTE. The decrease in overall accuracy for each class obfuscation again tracks with the count of samples within application classes~0, 2, and 3. However, as the obfuscated class is never considered for prediction by RF, we observe a lesser overall accuracy decrease as compared to scenario case~1.


\begin{figure}[!htb]
	\centering
	\begin{tabular}{cc}
	\adjustbox{scale=0.8}{
	\input figures/conf_2_a.tex
	}
	&
	\adjustbox{scale=0.8}{
	\input figures/conf_2_b.tex
	}
	\\
	(a) class $0 \longrightarrow 5$
	&
	(b) class $2 \longrightarrow 3$
	\\
	\\[-1.5ex]
	\multicolumn{2}{c}{\adjustbox{scale=0.8}{
	\input figures/conf_2_c.tex
	}}
	\\
	\multicolumn{2}{c}{(c) class $3 \longrightarrow 7$}
	\end{tabular}
	\caption{Confusion matrices for obfuscation scenario case~2} 
	\label{fig:case2}
\end{figure}

In scenario case~3, we train our RF classifier with the same obfuscated class in both the training dataset and the test dataset. We obfuscate only a small portion of the training data while still obfuscating all of the test data for each of class~0, 2, and 3. We experiment with the percentage of training data we obfuscate. This represents a hypothetical scenario where the obfuscation algorithm has been obfuscating network traffic long enough to pollute a small portion of a network traffic population. A defender then updates the classifier to include this small portion of obfuscated class data at training time, with increasing exposure to the obfuscated data over time. As our dataset is split into~80\% training data and~20\% test data, we decide to limit the training dataset exposure of obfuscated class data to~20\% of the total training dataset. We choose to decrement this value logarithmically with three total sub-scenarios representing~0.2\%, 2\%, and~20\% obfuscation exposure, expecting that with more exposure to the obfuscated class data, our classifier will adapt and outperform the obfuscation algorithm to correctly classify the obfuscated class in our test dataset. 

We find that~20\% exposure of our obfuscation algorithm to the RF training data is sufficient for RF to predict the disguised classes with high accuracy, defeating our obfuscation technique as shown in Table~\ref{tab:obacc}. Note that the overall accuracies reported for scenario case~3 are higher than our RF benchmark score of~92.2\%. We modify the test dataset in both scenario cases~1 
and~3, so the resulting accuracies of those scenarios cannot be compared to the results of prior work. Our results with lower exposure levels of~2\% and~0.2\% reveal a trend. Class~0 appears to be the most difficult class for our algorithm to obfuscate, while class~3 appears to be the easiest for our algorithm to obfuscate. Class~2 is somewhere in between, providing a loose correlation to our metric of statistical distance between classes and the performance of our obfuscation algorithm. We observe this trend in Table~\ref{tab:obacc} under ``Class Accuracy'' for ``Case~3.''

\begin{table}[!htb]
	\caption{Class accuracies for obfuscation scenarios}\label{tab:obacc}
	\centering
	\adjustbox{scale=0.9}{
		\begin{tabular}{c|c||c|c|c||c|c|c} \midrule\midrule
		\multirow{2}{*}{Case} & \multirow{2}{*}{Obfuscation} &  \multicolumn{3}{c||}{Overall Accuracy} & \multicolumn{3}{c}{Class Accuracy} \\
		    &  & (0,5) & (2,3) & (3,7) & 0 & 2 & 3 \\ 
		    \midrule
		    Original & --- & --- & --- & --- & 0.907 & 0.846 & 0.821 \\
		    1 & --- & 0.808 & 0.854 & 0.887 & 0.100 & 0.006 & 0.000 \\
		    2 & --- & 0.820 & 0.865 & 0.894 & 0.000 & 0.000 & 0.000 \\
		    3 & 20.0\% & 0.947 & 0.946 & 0.939 & 0.998 & 0.993 & 0.997 \\
		    3 & \zz2.0\% & 0.935 & 0.939 & 0.891 & 0.958 & 0.921 & 0.120 \\
		    3 & \zz0.2\% & 0.820 & 0.859 & 0.887 & 0.503 & 0.247 & 0.000 \\
		    \midrule\midrule
		\end{tabular}
	}
\end{table}


\section{Conclusion and Future Work\label{chap:conclusion}}

In this research, we classified the CIC-Darknet2020 network traffic samples and assessed SVM, RF, CNN, and 
the discriminator model of AC-GAN as our classifiers. We classified~4 traffic and~8 application classes while fine tuning the classifier hyperparameters. We experimented with different levels of SMOTE to assess class imbalance in the dataset and explored \hbox{2-D} representations of the traffic features for CNN and AC-GAN. We also approached the issue of darknet detection adversarially, from the perspective of an attacker hoping to confuse our best classifier. We demonstrated that we could effectively obfuscate application class traffic features. We then correlated the underlying statistics of the CIC-Darknet2020 dataset to the performance of this algorithm assuming some hypothetical scenarios for added realism.

\begin{figure}[!htb]
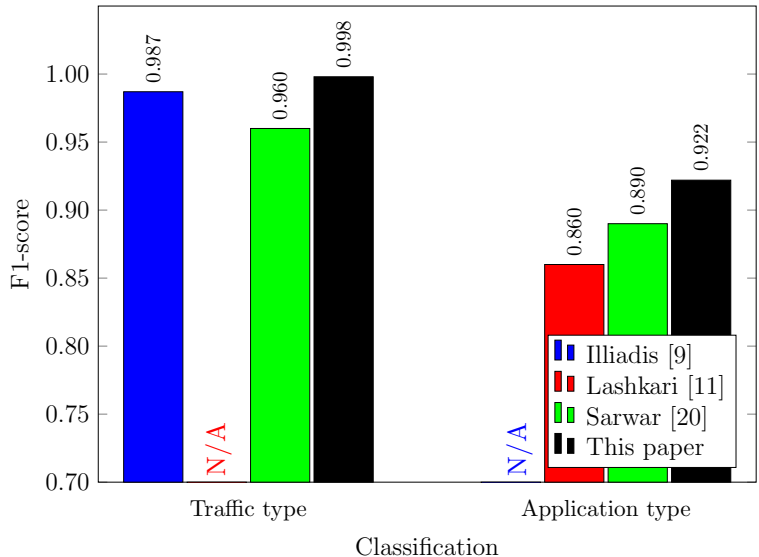

    \centering
    \input figures/compare.tex
    \caption{Classification accuracy compared to previous work}\label{fig:accComp}
\end{figure}

Among the tested machine learning classifiers, RF was found to be the most proficient at classifying darknet traffic for both traffic and application types. It yielded~99.9\% F1-score for traffic classification and~92.3\% F1-score for application classification, outperforming the state-of-the-art studies on CIC-Darknet2020~\cite{iliadis2021,cnnlstm2021}. Figure~\ref{fig:accComp}
provides a visual comparison of our best results with those of prior work.

Our research was limited by the availability of darknet traffic datasets. We selected the CIC-Darknet2020 dataset because it is frequently cited and publicly accessible, however the dataset suffers from a substantial imbalance. We attempted to compensate for this class imbalance by generating artificial samples with AC-GAN and SMOTE. The artificial SMOTE samples did not appear realistic for this dataset, which contains primarily non-linear feature values. Seeking to improve the quality of artificial samples, we assessed AC-GAN as a sample generator. However, our AC-GAN generated samples were not useful for data augmentation purposes. An approach future research might consider is to use clustering to group samples within a class, then train one GAN per cluster to generate samples. Other variations of GAN might also be better suited for multiclass sample generation and could generate more authentic samples.

We kept our obfuscation fairly basic, with the goal being to simply demonstrate that we could confuse our best classifier, with few restrictions imposed on the algorithm. Under realistic scenarios, it may be impossible to so easily modify features which define darknets such as Tor and VPN, but possible to obfuscate traffic features at the application layer such as those produced by CICFlowMeter analysis. We introduced a loose correlation to one statistical metric, an independent (Manhattan) sum of distances between DCPD across all sample features. We noted that~2 out of the~3 classes we chose to obfuscate were misclassified not as the intended classes, but with a majority of predictions distributed among other classes. This results from the fact that our obfuscation metric does not account for the statistical relationship between more than two classes, nor does it account for any dependency between the CIC-Darknet2020 feature values.

There is much remaining that could be done to extend the adversarial obfuscation analysis presented this paper. Real traffic features could be modified on live network traffic (e.g. changing IP addresses, ports, packet lengths or intervals), or select features could be prohibited from modification during obfuscation, which is likely to be a realistic constraint. An even larger task is to explore the dependency between features in order to anticipate counterattacks. One possible avenue that future research could take with the CIC-Darknet2020 dataset is to develop an obfuscation method to exploit RF feature importance, or the weights of a linear SVM. This might better correlate the relationship between classifier response and dataset statistics. We only tested our obfuscation method using our best-performing classifier. It would be interesting to explore how other classifiers respond to similar obfuscation techniques.

\bibliographystyle{plain}
\bibliography{references.bib}

\end{document}

%% file: preamble.tex
\usepackage{amsmath,amsthm, amsfonts, amssymb, amsxtra, amsopn}
\usepackage{pgfplots}
\pgfplotsset{compat=1.13}
\usepackage{pgfplotstable}
\usepackage{graphicx,grffile}
\usepackage{multirow}
\usepackage{booktabs}
\usepackage{listings}
\usepackage{cmap}
\usepackage{colortbl}
\usepackage{adjustbox}
\usepackage{epsfig}

\usepackage{algorithm} 
\usepackage{algpseudocode}

\usepackage[tableposition=top]{caption}
\usepackage{subcaption}
\usepackage{makecell} 

\PassOptionsToPackage{hyphens}{url}

\usepackage{hyperref}
\hypersetup{colorlinks=true,linkcolor=black,citecolor=black,urlcolor=blue,filecolor=black}
\hypersetup{pdfpagemode=UseNone,pdfstartview=}

\usepackage{enumitem}
\setlist[itemize]{noitemsep, topsep=0pt}

\advance\oddsidemargin by -0.2in
\advance\textwidth by 0.4in

\advance\topmargin by -0.25in
\advance\textheight by 0.5in

\long\def\symbolfootnotetext[#1]#2{\begingroup%
\def\thefootnote{\fnsymbol{footnote}}\footnotetext[#1]{#2}\endgroup}

\def\zz{\phantom{0}}

%% file: figures/conf_RF.tex
\begin{tikzpicture}[scale=0.75]
    \begin{axis}[
        width=10cm,
        height=10cm,
        xlabel={\large Predicted class},
        ylabel={\large Actual class},
	colormap={bluewhite}{color=(white) rgb255=(100,149,237)},
        xticklabels={
audio streaming,
browsing,
chat,
email,
file transfer,
P2P,
video streaming,
VOIP
        },
        xtick={0,...,7},
        xtick style={draw=none},
	xticklabel style={anchor=east,rotate=60,yshift=-5pt,font=\tt},
        yticklabels={
audio streaming,
browsing,
chat,
email,
file transfer,
P2P,
video streaming,
VOIP
        },
        ytick={0,...,7},
        ytick style={draw=none},
        enlargelimits=false,
        yticklabel style={font=\tt},
        colorbar,
        colorbar style={
            ytick={0.0,0.2,0.4,0.6,0.8,1.0},
            yticklabels={0.0,0.2,0.4,0.6,0.8,1.0},
            yticklabel={\pgfmathprintnumber\tick},
            yticklabel style={
            		/pgf/number format/fixed,
			/pgf/number format/precision=1}
        },
        point meta min=0.0,
        point meta max=1.0,
        nodes near coords={\pgfmathprintnumber\pgfplotspointmeta},
        nodes near coords black white/.style={
            small value/.style={
                yshift=-7pt,
                text=black,
                /pgf/number format/fixed,
                /pgf/number format/precision=3,
                /pgf/number format/zerofill=true,
                scale=0.9,
            },
            large value/.style={
                yshift=-7pt,
                text=white,
                /pgf/number format/fixed,
                /pgf/number format/precision=3,
                /pgf/number format/zerofill=true,
                scale=0.9,
            },
            every node near coord/.style={
                check for zero/.code={
                    \pgfmathfloatifflags{\pgfplotspointmeta}{0}{
                        \pgfkeys{/tikz/coordinate}
                    }{
                        \begingroup
                        \pgfkeys{/pgf/fpu}
                        \pgfmathparse{\pgfplotspointmeta<#1}
                        \global\let\result=\pgfmathresult
                        \endgroup
                        %
                        %
                        \pgfmathfloatcreate{1}{1.0}{0}
                        \let\ONE=\pgfmathresult
                        \ifx\result\ONE
                            \pgfkeysalso{/pgfplots/small value}
                        \else
                            \pgfkeysalso{/pgfplots/large value}
                        \fi
                    }
                },
                check for zero,
            },
        },
        nodes near coords black white=0.5,
    ]
        \addplot[
            matrix plot,
            mesh/cols=8,
            point meta=explicit,draw=gray
        ] table [meta=C] {
            x y C
0 0 0.907
1 0 0.003
2 0 0.003
3 0 0.000
4 0 0.004
5 0 0.000
6 0 0.082
7 0 0.001
0 1 0.006
1 1 0.940
2 1 0.003
3 1 0.000
4 1 0.030
5 1 0.001
6 1 0.021
7 1 0.000
0 2 0.003
1 2 0.001
2 2 0.846
3 2 0.107
4 2 0.013
5 2 0.000
6 2 0.002
7 2 0.029
0 3 0.001
1 3 0.001
2 3 0.165
3 3 0.821
4 3 0.006
5 3 0.000
6 3 0.002
7 3 0.003
0 4 0.009
1 4 0.051
2 4 0.012
3 4 0.003
4 4 0.899
5 4 0.000
6 4 0.023
7 4 0.003
0 5 0.001
1 5 0.001
2 5 0.000
3 5 0.000
4 5 0.001
5 5 0.996
6 5 0.000
7 5 0.000
0 6 0.158
1 6 0.026
2 6 0.008
3 6 0.002
4 6 0.040
5 6 0.000
6 6 0.760
7 6 0.006
0 7 0.003
1 7 0.001
2 7 0.171
3 7 0.022
4 7 0.016
5 7 0.000
6 7 0.005
7 7 0.782
         };
    \end{axis}
\end{tikzpicture}

%% file: figures/conf_1_a.tex
\begin{tikzpicture}[scale=0.75]
    \begin{axis}[
        width=10cm,
        height=10cm,
        xlabel={\large Predicted class},
        ylabel={\large Actual class},
	colormap={bluewhite}{color=(white) rgb255=(100,149,237)},
        xticklabels={
0,
1,
2,
3,
4,
5,
6,
7
        },
        xtick={0,...,7},
        xtick style={draw=none},
	xticklabel style={font=\tt},
        yticklabels={
0,
1,
2,
3,
4,
5,
6,
7
        },
        ytick={0,...,7},
        ytick style={draw=none},
        enlargelimits=false,
        yticklabel style={font=\tt},
        colorbar,
        colorbar style={
            ytick={0.0,0.2,0.4,0.6,0.8,1.0},
            yticklabels={0.0,0.2,0.4,0.6,0.8,1.0},
            yticklabel={\pgfmathprintnumber\tick},
            yticklabel style={
            		/pgf/number format/fixed,
			/pgf/number format/precision=1}
        },
        point meta min=0.0,
        point meta max=1.0,
        nodes near coords={\pgfmathprintnumber\pgfplotspointmeta},
        nodes near coords black white/.style={
            small value/.style={
                yshift=-7pt,
                text=black,
                /pgf/number format/fixed,
                /pgf/number format/precision=3,
                /pgf/number format/zerofill=true,
                scale=0.9,
            },
            large value/.style={
                yshift=-7pt,
                text=white,
                /pgf/number format/fixed,
                /pgf/number format/precision=3,
                /pgf/number format/zerofill=true,
                scale=0.9,
            },
            every node near coord/.style={
                check for zero/.code={
                    \pgfmathfloatifflags{\pgfplotspointmeta}{0}{
                        \pgfkeys{/tikz/coordinate}
                    }{
                        \begingroup
                        \pgfkeys{/pgf/fpu}
                        \pgfmathparse{\pgfplotspointmeta<#1}
                        \global\let\result=\pgfmathresult
                        \endgroup
                        %
                        %
                        \pgfmathfloatcreate{1}{1.0}{0}
                        \let\ONE=\pgfmathresult
                        \ifx\result\ONE
                            \pgfkeysalso{/pgfplots/small value}
                        \else
                            \pgfkeysalso{/pgfplots/large value}
                        \fi
                    }
                },
                check for zero,
            },
        },
        nodes near coords black white=0.5,
    ]
        \addplot[
            matrix plot,
            mesh/cols=8,
            point meta=explicit,draw=gray
        ] table [meta=C] {
            x y C
0 0 0.100
1 0 0.013
2 0 0.038
3 0 0.003
4 0 0.055
5 0 0.010
6 0 0.774
7 0 0.008
0 1 0.000
1 1 0.948
2 1 0.003
3 1 0.000
4 1 0.028
5 1 0.001
6 1 0.019
7 1 0.000
0 2 0.001
1 2 0.002
2 2 0.845
3 2 0.112
4 2 0.012
5 2 0.000
6 2 0.002
7 2 0.026
0 3 0.000
1 3 0.002
2 3 0.152
3 3 0.835
4 3 0.008
5 3 0.000
6 3 0.001
7 3 0.003
0 4 0.340
1 4 0.038
2 4 0.010
3 4 0.002
4 4 0.590
5 4 0.000
6 4 0.019
7 4 0.002
0 5 0.000
1 5 0.000
2 5 0.000
3 5 0.000
4 5 0.001
5 5 0.998
6 5 0.000
7 5 0.000
0 6 0.595
1 6 0.012
2 6 0.004
3 6 0.001
4 6 0.017
5 6 0.000
6 6 0.368
7 6 0.002
0 7 0.000
1 7 0.001
2 7 0.173
3 7 0.026
4 7 0.023
5 7 0.000
6 7 0.003
7 7 0.774
         };
    \end{axis}
\draw[black,thick] (5.8,7.825) circle(0.5);
\draw[black,dashed,thick] (0.55,7.825) circle(0.5);
\end{tikzpicture}

%% file: figures/conf_1_b.tex
\begin{tikzpicture}[scale=0.75]
    \begin{axis}[
        width=10cm,
        height=10cm,
        xlabel={\large Predicted class},
        ylabel={\large Actual class},
	colormap={bluewhite}{color=(white) rgb255=(100,149,237)},
        xticklabels={
0,
1,
2,
3,
4,
5,
6,
7
        },
        xtick={0,...,7},
        xtick style={draw=none},
	xticklabel style={font=\tt},
        yticklabels={
0,
1,
2,
3,
4,
5,
6,
7
        },
        ytick={0,...,7},
        ytick style={draw=none},
        enlargelimits=false,
        yticklabel style={font=\tt},
        colorbar,
        colorbar style={
            ytick={0.0,0.2,0.4,0.6,0.8,1.0},
            yticklabels={0.0,0.2,0.4,0.6,0.8,1.0},
            yticklabel={\pgfmathprintnumber\tick},
            yticklabel style={
            		/pgf/number format/fixed,
			/pgf/number format/precision=1}
        },
        point meta min=0.0,
        point meta max=1.0,
        nodes near coords={\pgfmathprintnumber\pgfplotspointmeta},
        nodes near coords black white/.style={
            small value/.style={
                yshift=-7pt,
                text=black,
                /pgf/number format/fixed,
                /pgf/number format/precision=3,
                /pgf/number format/zerofill=true,
                scale=0.9,
            },
            large value/.style={
                yshift=-7pt,
                text=white,
                /pgf/number format/fixed,
                /pgf/number format/precision=3,
                /pgf/number format/zerofill=true,
                scale=0.9,
            },
            every node near coord/.style={
                check for zero/.code={
                    \pgfmathfloatifflags{\pgfplotspointmeta}{0}{
                        \pgfkeys{/tikz/coordinate}
                    }{
                        \begingroup
                        \pgfkeys{/pgf/fpu}
                        \pgfmathparse{\pgfplotspointmeta<#1}
                        \global\let\result=\pgfmathresult
                        \endgroup
                        %
                        %
                        \pgfmathfloatcreate{1}{1.0}{0}
                        \let\ONE=\pgfmathresult
                        \ifx\result\ONE
                            \pgfkeysalso{/pgfplots/small value}
                        \else
                            \pgfkeysalso{/pgfplots/large value}
                        \fi
                    }
                },
                check for zero,
            },
        },
        nodes near coords black white=0.5,
    ]
        \addplot[
            matrix plot,
            mesh/cols=8,
            point meta=explicit,draw=gray
        ] table [meta=C] {
            x y C
0 0 0.898
1 0 0.001
2 0 0.005
3 0 0.000
4 0 0.006
5 0 0.001
6 0 0.086
7 0 0.001
0 1 0.007
1 1 0.945
2 1 0.000
3 1 0.000
4 1 0.028
5 1 0.001
6 1 0.019
7 1 0.000
0 2 0.020
1 2 0.014
2 2 0.006
3 2 0.707
4 2 0.073
5 2 0.000
6 2 0.014
7 2 0.166
0 3 0.002
1 3 0.002
2 3 0.000
3 3 0.982
4 3 0.009
5 3 0.000
6 3 0.001
7 3 0.004
0 4 0.006
1 4 0.032
2 4 0.447
3 4 0.002
4 4 0.496
5 4 0.000
6 4 0.016
7 4 0.002
0 5 0.001
1 5 0.000
2 5 0.000
3 5 0.000
4 5 0.001
5 5 0.998
6 5 0.000
7 5 0.000
0 6 0.118
1 6 0.019
2 6 0.236
3 6 0.002
4 6 0.028
5 6 0.000
6 6 0.592
7 6 0.004
0 7 0.000
1 7 0.002
2 7 0.000
3 7 0.032
4 7 0.027
5 7 0.000
6 7 0.003
7 7 0.936
         };
    \end{axis}
\draw[black,dashed,thick] (2.65,5.725) circle(0.5);
\draw[black,thick] (3.7,5.725) circle(0.5);
\end{tikzpicture}

%% file: figures/conf_1_c.tex
\begin{tikzpicture}[scale=0.75]
    \begin{axis}[
        width=10cm,
        height=10cm,
        xlabel={\large Predicted class},
        ylabel={\large Actual class},
	colormap={bluewhite}{color=(white) rgb255=(100,149,237)},
        xticklabels={
0,
1,
2,
3,
4,
5,
6,
7
        },
        xtick={0,...,7},
        xtick style={draw=none},
	xticklabel style={font=\tt},
        yticklabels={
0,
1,
2,
3,
4,
5,
6,
7
        },
        ytick={0,...,7},
        ytick style={draw=none},
        enlargelimits=false,
        yticklabel style={font=\tt},
        colorbar,
        colorbar style={
            ytick={0.0,0.2,0.4,0.6,0.8,1.0},
            yticklabels={0.0,0.2,0.4,0.6,0.8,1.0},
            yticklabel={\pgfmathprintnumber\tick},
            yticklabel style={
            		/pgf/number format/fixed,
			/pgf/number format/precision=1}
        },
        point meta min=0.0,
        point meta max=1.0,
        nodes near coords={\pgfmathprintnumber\pgfplotspointmeta},
        nodes near coords black white/.style={
            small value/.style={
                yshift=-7pt,
                text=black,
                /pgf/number format/fixed,
                /pgf/number format/precision=3,
                /pgf/number format/zerofill=true,
                scale=0.9,
            },
            large value/.style={
                yshift=-7pt,
                text=white,
                /pgf/number format/fixed,
                /pgf/number format/precision=3,
                /pgf/number format/zerofill=true,
                scale=0.9,
            },
            every node near coord/.style={
                check for zero/.code={
                    \pgfmathfloatifflags{\pgfplotspointmeta}{0}{
                        \pgfkeys{/tikz/coordinate}
                    }{
                        \begingroup
                        \pgfkeys{/pgf/fpu}
                        \pgfmathparse{\pgfplotspointmeta<#1}
                        \global\let\result=\pgfmathresult
                        \endgroup
                        %
                        %
                        \pgfmathfloatcreate{1}{1.0}{0}
                        \let\ONE=\pgfmathresult
                        \ifx\result\ONE
                            \pgfkeysalso{/pgfplots/small value}
                        \else
                            \pgfkeysalso{/pgfplots/large value}
                        \fi
                    }
                },
                check for zero,
            },
        },
        nodes near coords black white=0.5,
    ]
        \addplot[
            matrix plot,
            mesh/cols=8,
            point meta=explicit,draw=gray
        ] table [meta=C] {
            x y C
0 0 0.900
1 0 0.001
2 0 0.004
3 0 0.000
4 0 0.006
5 0 0.001
6 0 0.087
7 0 0.001
0 1 0.007
1 1 0.942
2 1 0.003
3 1 0.000
4 1 0.027
5 1 0.001
6 1 0.019
7 1 0.000
0 2 0.003
1 2 0.002
2 2 0.949
3 2 0.000
4 2 0.013
5 2 0.000
6 2 0.002
7 2 0.029
0 3 0.010
1 3 0.010
2 3 0.907
3 3 0.000
4 3 0.046
5 3 0.000
6 3 0.005
7 3 0.021
0 4 0.008
1 4 0.042
2 4 0.011
3 4 0.270
4 4 0.648
5 4 0.000
6 4 0.020
7 4 0.002
0 5 0.001
1 5 0.000
2 5 0.000
3 5 0.000
4 5 0.001
5 5 0.997
6 5 0.000
7 5 0.000
0 6 0.124
1 6 0.020
2 6 0.007
3 6 0.196
4 6 0.029
5 6 0.000
6 6 0.620
7 6 0.004
0 7 0.000
1 7 0.001
2 7 0.178
3 7 0.000
4 7 0.023
5 7 0.000
6 7 0.003
7 7 0.795
         };
    \end{axis}
\draw[black,dashed,thick] (3.7,4.675) circle(0.5);
\draw[black,thick] (7.9,4.675) circle(0.5);
\end{tikzpicture}

%% file: figures/conf_2_a.tex
\begin{tikzpicture}[scale=0.75]
    \begin{axis}[
        width=10cm,
        height=10cm,
        xlabel={\large Predicted class},
        ylabel={\large Actual class},
	colormap={bluewhite}{color=(white) rgb255=(100,149,237)},
        xticklabels={
0,
1,
2,
3,
4,
5,
6,
7
        },
        xtick={0,...,7},
        xtick style={draw=none},
	xticklabel style={font=\tt},
        yticklabels={
0,
1,
2,
3,
4,
5,
6,
7
        },
        ytick={0,...,7},
        ytick style={draw=none},
        enlargelimits=false,
        yticklabel style={font=\tt},
        colorbar,
        colorbar style={
            ytick={0.0,0.2,0.4,0.6,0.8,1.0},
            yticklabels={0.0,0.2,0.4,0.6,0.8,1.0},
            yticklabel={\pgfmathprintnumber\tick},
            yticklabel style={
            		/pgf/number format/fixed,
			/pgf/number format/precision=1}
        },
        point meta min=0.0,
        point meta max=1.0,
        nodes near coords={\pgfmathprintnumber\pgfplotspointmeta},
        nodes near coords black white/.style={
            small value/.style={
                yshift=-7pt,
                text=black,
                /pgf/number format/fixed,
                /pgf/number format/precision=3,
                /pgf/number format/zerofill=true,
                scale=0.9,
            },
            large value/.style={
                yshift=-7pt,
                text=white,
                /pgf/number format/fixed,
                /pgf/number format/precision=3,
                /pgf/number format/zerofill=true,
                scale=0.9,
            },
            every node near coord/.style={
                check for zero/.code={
                    \pgfmathfloatifflags{\pgfplotspointmeta}{0}{
                        \pgfkeys{/tikz/coordinate}
                    }{
                        \begingroup
                        \pgfkeys{/pgf/fpu}
                        \pgfmathparse{\pgfplotspointmeta<#1}
                        \global\let\result=\pgfmathresult
                        \endgroup
                        %
                        %
                        \pgfmathfloatcreate{1}{1.0}{0}
                        \let\ONE=\pgfmathresult
                        \ifx\result\ONE
                            \pgfkeysalso{/pgfplots/small value}
                        \else
                            \pgfkeysalso{/pgfplots/large value}
                        \fi
                    }
                },
                check for zero,
            },
        },
        nodes near coords black white=0.5,
    ]
        \addplot[
            matrix plot,
            mesh/cols=8,
            point meta=explicit,draw=gray
        ] table [meta=C] {
            x y C
0 0 0.000
1 0 0.000
2 0 0.000
3 0 0.000
4 0 0.000
5 0 0.000
6 0 0.000
7 0 0.000
0 1 0.013
1 1 0.936
2 1 0.003
3 1 0.000
4 1 0.027
5 1 0.002
6 1 0.018
7 1 0.000
0 2 0.359
1 2 0.002
2 2 0.542
3 2 0.070
4 2 0.008
5 2 0.000
6 2 0.002
7 2 0.017
0 3 0.392
1 3 0.001
2 3 0.089
3 3 0.510
4 3 0.005
5 3 0.000
6 3 0.002
7 3 0.002
0 4 0.174
1 4 0.047
2 4 0.012
3 4 0.002
4 4 0.739
5 4 0.000
6 4 0.023
7 4 0.003
0 5 0.004
1 5 0.000
2 5 0.001
3 5 0.000
4 5 0.001
5 5 0.994
6 5 0.000
7 5 0.000
0 6 0.334
1 6 0.015
2 6 0.008
3 6 0.002
4 6 0.028
5 6 0.000
6 6 0.610
7 6 0.003
0 7 0.026
1 7 0.001
2 7 0.170
3 7 0.028
4 7 0.020
5 7 0.000
6 7 0.005
7 7 0.751
         };
    \end{axis}
\draw[black,thick] (5.8,7.825) circle(0.5);
\draw[black,dashed,thick] (0.55,7.825) circle(0.5);
\end{tikzpicture}

%% file: figures/conf_2_b.tex
\begin{tikzpicture}[scale=0.75]
    \begin{axis}[
        width=10cm,
        height=10cm,
        xlabel={\large Predicted class},
        ylabel={\large Actual class},
	colormap={bluewhite}{color=(white) rgb255=(100,149,237)},
        xticklabels={
0,
1,
2,
3,
4,
5,
6,
7
        },
        xtick={0,...,7},
        xtick style={draw=none},
	xticklabel style={font=\tt},
        yticklabels={
0,
1,
2,
3,
4,
5,
6,
7
        },
        ytick={0,...,7},
        ytick style={draw=none},
        enlargelimits=false,
        yticklabel style={font=\tt},
        colorbar,
        colorbar style={
            ytick={0.0,0.2,0.4,0.6,0.8,1.0},
            yticklabels={0.0,0.2,0.4,0.6,0.8,1.0},
            yticklabel={\pgfmathprintnumber\tick},
            yticklabel style={
            		/pgf/number format/fixed,
			/pgf/number format/precision=1}
        },
        point meta min=0.0,
        point meta max=1.0,
        nodes near coords={\pgfmathprintnumber\pgfplotspointmeta},
        nodes near coords black white/.style={
            small value/.style={
                yshift=-7pt,
                text=black,
                /pgf/number format/fixed,
                /pgf/number format/precision=3,
                /pgf/number format/zerofill=true,
                scale=0.9,
            },
            large value/.style={
                yshift=-7pt,
                text=white,
                /pgf/number format/fixed,
                /pgf/number format/precision=3,
                /pgf/number format/zerofill=true,
                scale=0.9,
            },
            every node near coord/.style={
                check for zero/.code={
                    \pgfmathfloatifflags{\pgfplotspointmeta}{0}{
                        \pgfkeys{/tikz/coordinate}
                    }{
                        \begingroup
                        \pgfkeys{/pgf/fpu}
                        \pgfmathparse{\pgfplotspointmeta<#1}
                        \global\let\result=\pgfmathresult
                        \endgroup
                        %
                        %
                        \pgfmathfloatcreate{1}{1.0}{0}
                        \let\ONE=\pgfmathresult
                        \ifx\result\ONE
                            \pgfkeysalso{/pgfplots/small value}
                        \else
                            \pgfkeysalso{/pgfplots/large value}
                        \fi
                    }
                },
                check for zero,
            },
        },
        nodes near coords black white=0.5,
    ]
        \addplot[
            matrix plot,
            mesh/cols=8,
            point meta=explicit,draw=gray
        ] table [meta=C] {
            x y C
0 0 0.883
1 0 0.002
2 0 0.025
3 0 0.000
4 0 0.006
5 0 0.001
6 0 0.082
7 0 0.001
0 1 0.007
1 1 0.941
2 1 0.004
3 1 0.000
4 1 0.027
5 1 0.001
6 1 0.019
7 1 0.000
0 2 0.000
1 2 0.000
2 2 0.000
3 2 0.000
4 2 0.000
5 2 0.000
6 2 0.000
7 2 0.000
0 3 0.001
1 3 0.001
2 3 0.486
3 3 0.497
4 3 0.005
5 3 0.000
6 3 0.000
7 3 0.009
0 4 0.010
1 4 0.045
2 4 0.196
3 4 0.002
4 4 0.719
5 4 0.000
6 4 0.026
7 4 0.003
0 5 0.001
1 5 0.000
2 5 0.001
3 5 0.000
4 5 0.001
5 5 0.997
6 5 0.000
7 5 0.000
0 6 0.142
1 6 0.023
2 6 0.058
3 6 0.003
4 6 0.038
5 6 0.000
6 6 0.732
7 6 0.006
0 7 0.000
1 7 0.001
2 7 0.335
3 7 0.027
4 7 0.022
5 7 0.000
6 7 0.004
7 7 0.612
         };
    \end{axis}
\draw[black,dashed,thick] (2.65,5.725) circle(0.5);
\draw[black,thick] (3.7,5.725) circle(0.5);
\end{tikzpicture}

%% file: figures/conf_2_c.tex
\begin{tikzpicture}[scale=0.75]
    \begin{axis}[
        width=10cm,
        height=10cm,
        xlabel={\large Predicted class},
        ylabel={\large Actual class},
	colormap={bluewhite}{color=(white) rgb255=(100,149,237)},
        xticklabels={
0,
1,
2,
3,
4,
5,
6,
7
        },
        xtick={0,...,7},
        xtick style={draw=none},
	xticklabel style={font=\tt},
        yticklabels={
0,
1,
2,
3,
4,
5,
6,
7
        },
        ytick={0,...,7},
        ytick style={draw=none},
        enlargelimits=false,
        yticklabel style={font=\tt},
        colorbar,
        colorbar style={
            ytick={0.0,0.2,0.4,0.6,0.8,1.0},
            yticklabels={0.0,0.2,0.4,0.6,0.8,1.0},
            yticklabel={\pgfmathprintnumber\tick},
            yticklabel style={
            		/pgf/number format/fixed,
			/pgf/number format/precision=1}
        },
        point meta min=0.0,
        point meta max=1.0,
        nodes near coords={\pgfmathprintnumber\pgfplotspointmeta},
        nodes near coords black white/.style={
            small value/.style={
                yshift=-7pt,
                text=black,
                /pgf/number format/fixed,
                /pgf/number format/precision=3,
                /pgf/number format/zerofill=true,
                scale=0.9,
            },
            large value/.style={
                yshift=-7pt,
                text=white,
                /pgf/number format/fixed,
                /pgf/number format/precision=3,
                /pgf/number format/zerofill=true,
                scale=0.9,
            },
            every node near coord/.style={
                check for zero/.code={
                    \pgfmathfloatifflags{\pgfplotspointmeta}{0}{
                        \pgfkeys{/tikz/coordinate}
                    }{
                        \begingroup
                        \pgfkeys{/pgf/fpu}
                        \pgfmathparse{\pgfplotspointmeta<#1}
                        \global\let\result=\pgfmathresult
                        \endgroup
                        %
                        %
                        \pgfmathfloatcreate{1}{1.0}{0}
                        \let\ONE=\pgfmathresult
                        \ifx\result\ONE
                            \pgfkeysalso{/pgfplots/small value}
                        \else
                            \pgfkeysalso{/pgfplots/large value}
                        \fi
                    }
                },
                check for zero,
            },
        },
        nodes near coords black white=0.5,
    ]
        \addplot[
            matrix plot,
            mesh/cols=8,
            point meta=explicit,draw=gray
        ] table [meta=C] {
            x y C
0 0 0.890
1 0 0.003
2 0 0.004
3 0 0.010
4 0 0.006
5 0 0.002
6 0 0.085
7 0 0.001
0 1 0.007
1 1 0.942
2 1 0.003
3 1 0.002
4 1 0.026
5 1 0.002
6 1 0.019
7 1 0.000
0 2 0.002
1 2 0.003
2 2 0.667
3 2 0.296
4 2 0.009
5 2 0.000
6 2 0.002
7 2 0.022
0 3 0.000
1 3 0.000
2 3 0.000
3 3 0.000
4 3 0.000
5 3 0.000
6 3 0.000
7 3 0.000
0 4 0.012
1 4 0.049
2 4 0.012
3 4 0.098
4 4 0.800
5 4 0.000
6 4 0.026
7 4 0.003
0 5 0.001
1 5 0.000
2 5 0.000
3 5 0.001
4 5 0.001
5 5 0.997
6 5 0.000
7 5 0.000
0 6 0.148
1 6 0.021
2 6 0.009
3 6 0.017
4 6 0.036
5 6 0.000
6 6 0.764
7 6 0.005
0 7 0.000
1 7 0.001
2 7 0.180
3 7 0.028
4 7 0.020
5 7 0.000
6 7 0.004
7 7 0.768
         };
    \end{axis}
\draw[black,dashed,thick] (3.7,4.675) circle(0.5);
\draw[black,thick] (7.9,4.675) circle(0.5);
\end{tikzpicture}

%% file: figures/compare.tex
\begin{tikzpicture}[scale=0.8, every node/.style={scale=1.0}]
    \begin{axis}[
        width  = 0.85*\textwidth,
        height = 9.5cm,
        ymin=0.70,ymax=1.05,
        ytick={0.70,0.75,0.80,0.85,0.90,0.95,1.00},
        major x tick style = transparent,
        ybar=5*\pgflinewidth,
        bar width=28.0pt,
        xlabel = {Classification},
        ylabel = {F1-score},
        xtick={1,2},
        xticklabels={
        		Traffic type,
		Application type},
	y tick label style={
    		/pgf/number format/.cd,
   		fixed,
   		fixed zerofill,
    		precision=2},
        x tick label style={
		font=\small,
		},
        nodes near coords,
        every node near coord/.append style={rotate=90, 
        								   anchor=west,
								   font=\footnotesize,
								   /pgf/number format/.cd,
								   	fixed zerofill,
									precision=3
								   },
        enlarge x limits=0.42,
        legend cell align=left,
        legend pos=south east,
    ]
\addplot[fill=blue,opacity=1.00] 
coordinates {
(1, 0.987)
(2, 0.000)
};
\addplot[fill=red,opacity=1.00] 
coordinates {
(1, 0.000)
(2, 0.860)
};
\addplot[fill=green,opacity=1.00] 
coordinates {
(1, 0.960)
(2, 0.890)
};
\addplot[fill=black,opacity=1.00] 
coordinates {
(1, 0.998)
(2, 0.922)
};
\legend{\hspace*{0.05in}Illiadis~\cite{iliadis2021},
	\hspace*{0.05in}Lashkari~\cite{tornontor2017},
	\hspace*{0.05in}Sarwar~\cite{cnnlstm2021},
	\hspace*{0.05in}This paper}
\end{axis}
\node[rotate=90,color=red] at (2,0.55) {\footnotesize N/A};
\node[rotate=90,color=blue] at (7,0.55) {\footnotesize N/A};
\end{tikzpicture}